\documentclass{article}
\usepackage{fontspec}
\usepackage{float}
\usepackage{arxiv}

\usepackage{hyperref}       
\usepackage{url}            
\usepackage{booktabs}       
\usepackage{amsfonts}       
\usepackage{amsmath}
\usepackage{amssymb}
\usepackage{natbib}
\usepackage{nicefrac}       
\usepackage{microtype}      
\usepackage{lipsum}
\usepackage{graphicx}
\usepackage{subcaption}
\usepackage{algorithm}
\usepackage{algorithmic}
\usepackage{enumitem}
\usepackage{makecell}
\usepackage{xcolor}
\usepackage{lettrine}
\usepackage{array}  
\usepackage{multirow}
\usepackage{colortbl}
\usepackage{arydshln}
\usepackage{threeparttable}
\usepackage{varwidth}
\usepackage{listings}
\usepackage[bottom]{footmisc}
\usepackage[most]{tcolorbox}
\tcbuselibrary{listings}
\tcbuselibrary{breakable}
\usepackage{inconsolata}
\usepackage{etoolbox}
\usepackage{microtype}
\UseMicrotypeSet[protrusion]{basicmath}

\DeclareMathOperator*{\argmax}{arg\,max}

\usepackage[T1]{fontenc}
\usepackage{tgtermes}           
\usepackage[scale=0.9]{tgheros} 
\usepackage{tgcursor}            

\definecolor{mygray}{gray}{0.9}  
\usepackage{arydshln}

\setlist{
    topsep=3pt,       
    itemsep=3pt,      
    parsep=0pt,       
    leftmargin=1.5em, 
    partopsep=0pt
}
\usepackage{tcolorbox}
\tcbuselibrary{skins, breakable, listingsutf8}

\lstdefinestyle{sharegpt_lst_style}{
    basicstyle=\linespread{1.15}\ttfamily\small\selectfont\microtypesetup{activate=false},
    columns=fullflexible,
    breaklines=true,
    breakindent=0pt,
    showstringspaces=false,
    xleftmargin=0pt,
    xrightmargin=0pt,
    aboveskip=0pt,
    belowskip=0pt,
    upquote=false, 
    tabsize=2,
    literate={—}{{--}}1 {“}{{``}}1 {”}{{''}}1 {’}{{'}}1
}

\newtcbinputlisting{\ExampleBox}[2]{
    listing only,
    listing file={#2},
    listing options={style=sharegpt_lst_style},
    title={#1},
    enhanced,
    breakable,
    colback=gray!10,
    colframe=black,
    grow to left by=0pt,    
    grow to right by=0pt,
    top=0pt,                
    bottom=0pt,
    left=0pt,           
    right=0pt,
    boxsep=1mm,
    colbacktitle=black,
    coltitle=white,
    boxsep=1mm,
    fonttitle=\bfseries\large\sffamily,
    boxrule=1pt,
    arc=5pt,
    toptitle=3pt,
    bottomtitle=3pt,
    sharp corners=south,
    extras first={
        sharp corners=south,
        bottomrule=1pt
    },
    extras middle={
        sharp corners,
        toprule=1pt,
        bottomrule=1pt
    },
    extras last={
        sharp corners=north,
        toprule=1pt
    }
}

\title{\raisebox{-0.4em}{\includegraphics[height=1.5em]{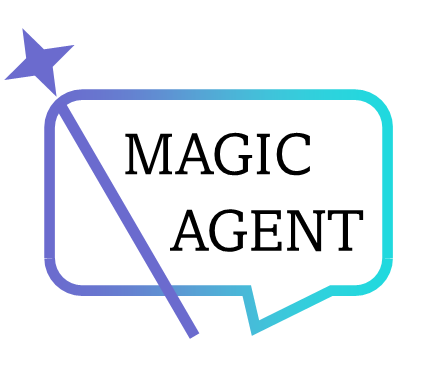}} MagicAgent: Towards Generalized Agent Planning}

\author{
  \renewcommand{\arraystretch}{1.5} 
  \begin{tabular}{c@{\hspace{.65cm}}c@{\hspace{.65cm}}c@{\hspace{.65cm}}c@{\hspace{.65cm}}c}
    Xuhui Ren \textsuperscript{1} &
    Shaokang Dong \textsuperscript{1} &
    Chen Yang \textsuperscript{1}  &
    Qing Gao \textsuperscript{1}  &
    Yunbin Zhao  \textsuperscript{1}  \\
    Yongsheng Liu \textsuperscript{1}  &
    Xinwei Geng \textsuperscript{1} &
    Xiang Li \textsuperscript{1} &
    Demei Yan \textsuperscript{1} &
    Yanqing Li \textsuperscript{1}  \\
    Chenhao Huang \textsuperscript{2} &
    Dingwei Zhu \textsuperscript{2} &
    Junjie Ye \textsuperscript{2} &
    Boxuan Yue \textsuperscript{1} &
    Yingnan Fu \textsuperscript{1} \\
    Mengzhe Lv \textsuperscript{1} &
    Zezeng Feng \textsuperscript{1} &
    Boshen Zhou \textsuperscript{1}  &
    Bocheng Wang \textsuperscript{1} &
    Xuanjing Huang \textsuperscript{2} \\
    \multicolumn{5}{c}{
    Yu-Gang Jiang \textsuperscript{2} \quad \quad \quad
    Tao Gui \textsuperscript{2} \footnotemark[1] \quad \quad \quad
    Qi Zhang \textsuperscript{2} \footnotemark[1] \quad \quad \quad
    Yunke Zhang \textsuperscript{1} \thanks{Corresponding authors: Tao Gui (tgui@fudan.edu.cn), Qi Zhang (qz@fudan.edu.cn), Yunke Zhang (zhangyunke@honor.com). The MagicAgent series models will be released soon.}
    }
  \end{tabular}
  \vspace{.1cm}
  \\
  \textit{\textsuperscript{1} Honor Device Co., Ltd} \vspace{.1cm}  \\
  \textit{\textsuperscript{2} Fudan University} \vspace{.2cm}
}

\begin{document}
\maketitle

\begin{abstract}
The evolution of Large Language Models (LLMs) from passive text processors to autonomous agents has established planning as a core component of modern intelligence. However, achieving generalized planning remains elusive, not only by the scarcity of high-quality interaction data but also by inherent conflicts across heterogeneous planning tasks. These challenges result in models that excel at isolated tasks yet struggle to generalize, while existing multi-task training attempts suffer from gradient interference. In this paper, we present \textbf{MagicAgent}, a series of foundation models specifically designed for generalized agent planning. We introduce a lightweight and scalable synthetic data framework that generates high-quality trajectories across diverse planning tasks, including hierarchical task decomposition, tool-augmented planning, multi-constraint scheduling, procedural logic orchestration, and long-horizon tool execution. To mitigate training conflicts, we propose a two-stage training paradigm comprising supervised fine-tuning followed by multi-objective reinforcement learning over both static datasets and dynamic environments. Empirical results show that MagicAgent-32B and MagicAgent-30B-A3B achieve superior performance across diverse open-source benchmarks (\emph{e.g.}, $75.1\%$ on Worfbench and $86.9\%$ on BFCL-v3), as well as strong results on our in-house MagicEval benchmarks, substantially outperforming existing sub-100B models and surpassing leading ultra-scale models, including GPT-5.2, Kimi-K2 and GLM-4.7.

\end{abstract}
\vspace{-3mm}

\begin{figure}[!h]
  \centering
  \includegraphics[scale=0.445]{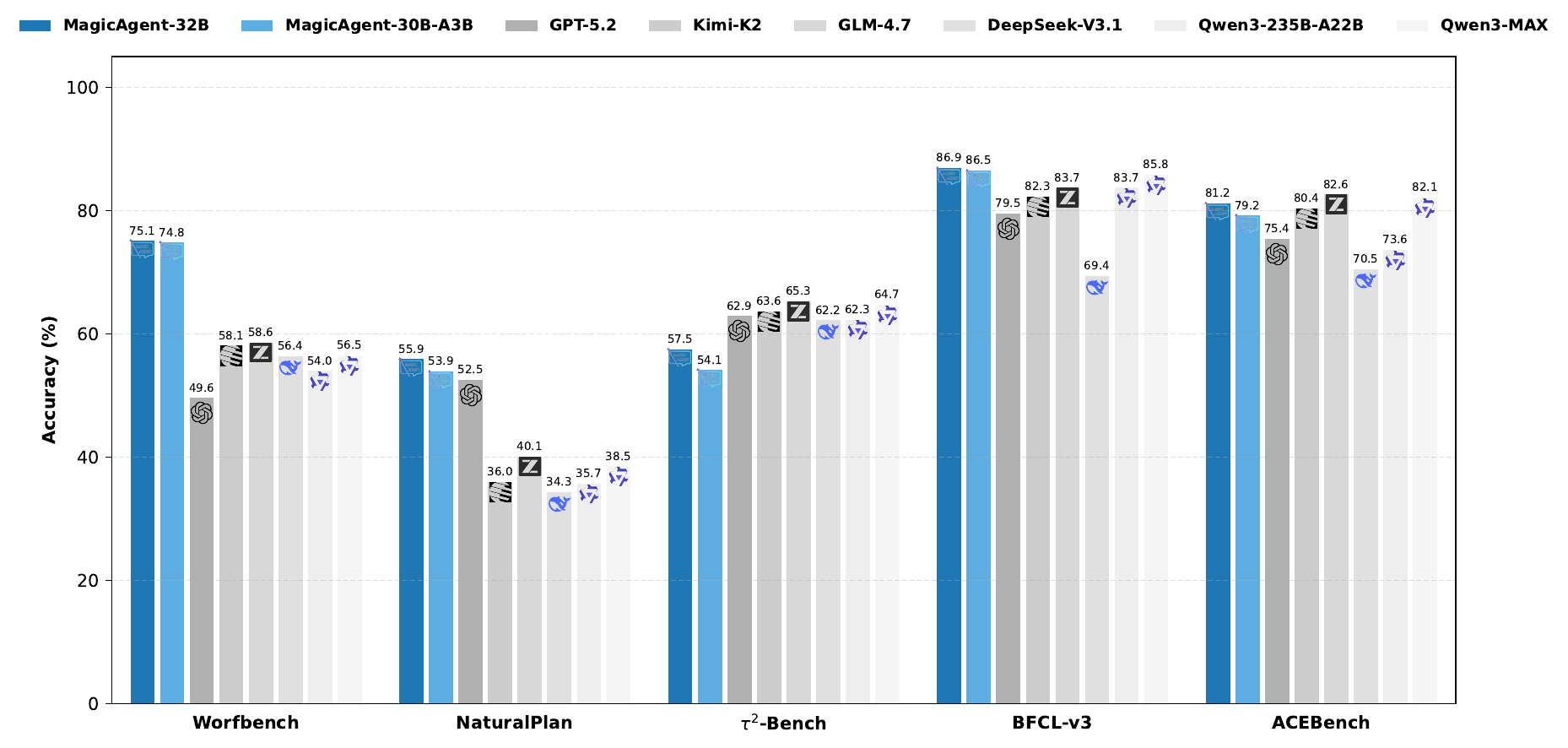}
  \caption{Benchmark performance of MagicAgent.}
  \label{fig:1}
\end{figure}
\section{INTRODUCTION}
The emergence of Large Language Models (LLMs) has catalyzed a paradigm shift across the Natural Language Processing (NLP) landscape, demonstrating unprecedented capabilities in language comprehension and generation \citep{CaoZhao2025}. These models have fundamentally transformed the nature of human-AI interaction, evolving from passive text processors \citep{openai2025gpt52, Qwen3} to autonomous agents capable of addressing complex, real-world challenges \citep{agents_survey, agentgym, webrl}. From autonomous website navigation and embodied intelligence to daily scheduling and virtual assistance \citep{zheng2024natural, Tau2bench}, LLMs are increasingly becoming central to modern computational ecosystems and large-scale intelligent agent systems.

At the core of agentic intelligence lies planning-a fundamental cognitive function that enables and orchestrates task execution. Effective planning empowers an agent to decompose complex objectives into a series of actionable sub-tasks and schedule their execution in a logical order, thereby significantly enhancing task completion rates and operational efficiency \citep{agents_survey}. To build a sophisticated planning agent, the coordinated integration of environmental perception, rigorous logical reasoning, and sequential decision-making is required. As LLMs transition to proactive decision-making agents, planning proficiency becomes a key determinant of their practical utility in diverse application scenarios. However, achieving generalized planning capabilities remains a substantial challenge. Historically, both industry and academia have relied on bespoke systems tailored to specific tasks through expert-crafted rules \citep{Human_agent, meta_gpt}. While these systems often achieve high precision within their narrow domains, they are inherently brittle; their reliance on rigid, predefined rules renders them incapable of generalizing to unseen scenarios and prohibitively expensive to scale across the complexity of real-world applications \citep{ALFWORLD, scienceworld, webshop, appworld_rl}. Modern LLMs, however, exhibit strong potential for broader generalization through prompting strategies, yet they remain constrained by the intrinsic limitations of the foundation model and the finite knowledge scope characteristic of static pretraining corpora.

Researchers have increasingly explored agent-centric model optimization to support diverse task-solving requirements. This paradigm typically involves fine-tuning models on task-specific trajectories to facilitate complex objective decomposition and execution. However, the development of robust, general-purpose agent planning is currently hindered by three primary limitations in existing methodologies: (1) \textbf{Data Scarcity and Generation Overhead}: A fundamental bottleneck lies in the scarcity of high-quality training data. Prior synthetic data generation approaches have largely relied on simulated human-agent interactions within virtual environments—a process that is both labor-intensive and cost-prohibitive \citep{environment_data, agents_survey}. (2) \textbf{Fragmented Capabilities in Bespoke Systems}: While progress has been made in tool-based planning, the broader landscape of general-purpose agent planning remains under-explored \citep{wangetal2025oolflow, yuaneta2025asytool}. Existing bespoke systems frequently focus on isolated capabilities; such single-point optimizations inherently lack the generalization ability and scalability required to establish a unified planning foundation. (3) \textbf{Optimization Instability in Multi-Task Learning}: Naive aggregation of multi-task planning data frequently induces a ``\textbf{seesaw effect}''. Competing optimization objectives across heterogeneous tasks impede stable joint convergence \citep{multi_task_survey, lora_multi_task, multi_task_rl1, multi_task_rl2}. In such scenarios, performance gains in one domain often trigger unintended degradation in others, further challenging the scalability of current agentic models.

We address these challenges by introducing a unified framework for training agent models that achieve superior performance across diverse planning scenarios. Recognizing data scarcity as a primary bottleneck, we develop a suite of lightweight synthetic data generation techniques spanning multiple planning dimensions, including task decomposition, tool-augmented planning, multi-constraint scheduling, and logical orchestration. Notably, we tackle complex tool chaining in long-horizon scenarios through a targeted trajectory generation framework, substantially improving the model's operational memory and execution stability. To mitigate the ``seesaw effect'', we propose a two-stage training paradigm that integrates supervised fine-tuning (SFT) with reinforcement learning (RL). In the SFT stage, we explicitly emphasize sample balance and diversity, enabling the coherent integration of heterogeneous planning datasets. Building on this foundation, the RL stage employs a unified multi-objective reward function to further enhance generalization on static datasets. Subsequently, we articulate the exploration-exploitation trade-off, thereby improving generalization in dynamic environments characterized by sparse rewards. Our framework supports both dense and mixture-of-experts (MoEs) architectures. Accordingly, we introduce a robust optimization strategy that stabilizes MoE training for agentic tasks and effectively alleviates expert load imbalance in multi-task settings.

Our \textbf{MagicAgent-32B} and \textbf{MagicAgent-30B-A3B} models achieve the accuracy of $80.3\%$ on Worfbench $F_1$ Chain, $69.9\%$ on Worfbench $F_1$ Graph, $55.9\%$ on NaturalPlan, $57.5\%$ on $\tau^2$-Bench, $86.9\%$ on BFCL-v3, and $81.2\%$ on ACEBench, as shown in Figure \ref{fig:1}, substantially outperforming existing sub-100B models on complex planning tasks, as well as several leading closed-source models, including GPT and DeepSeek. Additional evaluations on our in-house MagicEval-Plan and MagicEval-Tool benchmarks report as $98.0\%$ step accuracy, $95.9\%$ embedding semantic accuracy, $91.2\%$ LLM-as-a-Judge accuracy, $97.7\%$ tool name accuracy, and $87.3\%$ tool argument accuracy under general scenarios, further demonstrating the consistent and substantial performance advantage of MagicAgent.

The primary contributions of this work are summarized as follows:
\begin{itemize}
    \item \textbf{A Scalable Synthetic Data Framework}: A suite of lightweight methodologies is introduced to synthesize high-quality agentic data. By mitigating the efficiency bottlenecks of traditional sandbox-based data acquisition and the limitations of single-point optimization, this approach provides a cost-effective solution for generating data.
    
    \item \textbf{Two-Stage Multi-Task Optimization}: A novel training paradigm integrating SFT with offline and online RL is proposed to address the ``seesaw effect'' in multi-task learning, enabling synergistic utilization of heterogeneous static datasets while explicitly modeling the exploration-exploitation trade-off in online environments.
    
    \item \textbf{Load-Balanced MoE for Agents}: An optimization strategy is developed specifically for MoE architectures in agent planning, effectively overcoming expert load imbalance and ensuring stable training and strong performance in multi-task agentic scenarios.
    
    \item \textbf{State-of-the-art (SOTA) Performance}: The MagicAgent series is released, demonstrating that the 32B model establishes a new benchmark among sub-100B models. The results indicate that our planning-centric training can effectively enhance the performance of existing foundation models.
    
\end{itemize}

\section{RELATED WORK}
The development of artificial intelligence systems is rapidly transitioning from static, task-specific models to dynamic, agent-based architectures capable of generalizing across diverse applications. Driven by advancements in large-scale pretraining, supervised fine-tuning (SFT), and reinforcement learning (RL), LLMs have demonstrated remarkable capabilities in planning, reasoning, and natural language understanding. In this section, we review the primary approaches related to our work.

\subsection{Large Language Agent Models}
The advent of LLMs has positioned them at the forefront of artificial intelligence research, providing a robust foundation for the development of LLM-centric autonomous agents \citep{durante2024interactiveagentfoundationmodel, zhangetal2025xlam, durante2024agentaisurveyinghorizons, li2024personalllmagentsinsights,zhang2025step}. Numerous efforts have leveraged LLMs to construct agents tailored for specific applications, including tool utilization \citep{qiaoetal2024autoact, wangetal2025oolflow, yuaneta2025asytool, 1011456267723657828}, embodied planning \citep{react, wu2023embodiedtaskplanninglarge, Song023CCV}, web browsing \citep{he2024webvoyagerbuildingendtoendweb, 101145369641714842, wei2025webagentiningwebagents}, and software engineering \citep{lital25detree, dong2025surveynllmbased}. Despite these advancements, existing agent methodologies predominantly focus on isolated domains planning tasks, thereby lacking broader generalization capabilities.

A fundamental bottleneck hindering the generalization of large language agent models is the scarcity of high-quality agentic training data. To address this, several approaches explore simulating actual execution environments to collect behavioral trajectories for training robust agent models \citep{li2025chainofagentsendtoendagentfoundation, qiao2025benchmarking, li2025websailornavigatingsuperhumanreasoning}. To construct a large-scale tool-learning dataset, \citet{qin2023toolllmfacilitatin} collect thousands of real-world RESTful APIs to prompt a teacher model for diverse user instructions, and subsequently annotate successful multi-step execution trajectories by interacting with the real APIs via a depth-first search-based decision tree algorithm. Similarly, TaskCraft \citep{shi2025taskcraftautomatedgenerationagentic} utilizes depth-based and width-based extensions of atomic tasks to create verifiable trajectories. Furthermore, \citet{fang20agenticinenvironment} propose a scalable framework that automatically constructs fully simulated, database-backed environments from real-world API dependency graphs. Their method synthesizes multi-turn trajectories by executing logical tool sequences via random walks, retrospectively formulating complex user intents. However, these data generation paradigms remain labor- and time-intensive. Moreover, they primarily target specific planning tasks, resulting in a lack of diversity for generalized agent planning.

\subsection{Model Training Paradigms for Agents}

\textbf{SFT} constitutes a foundational stage in the adaptation of LLMs to agentic settings, enabling alignment with both general instructions and agent-oriented task distributions \citep{zhou2024enhancing}. The majority of existing studies adopt full-parameter SFT, leveraging carefully curated instruction corpora, interaction trajectories, or behavioral demonstrations to specialize pre-trained models for downstream agent tasks, as exemplified by AgentTuning \citep{Agenttuning}, Agent-FLAN \citep{Agentflan}, and AgentBank \citep{Agentbank}. To alleviate the substantial computational overhead associated with full fine-tuning, Parameter-efficient Fine-tuning (PEFT) techniques, including LoRA \citep{LoRA} and QLoRA \citep{QLoRA}, have been widely explored. These methods restrict optimization to a small subset of trainable parameters while freezing the backbone model, thereby achieving performance comparable to full-parameter tuning at a fraction of the cost \citep{Re-ReST,Fireact,SMART}. In addition to these general paradigms, recent work has proposed customized SFT formulations that incorporate task-aware optimization objectives, such as regularization-augmented loss functions \citep{Atm} or iterative refinement over trajectory data \citep{Interactive_evolution}, in order to improve the trade-off between task specialization and generalization. Taken together, SFT-based approaches offer a versatile and effective framework for tailoring LLMs to a wide spectrum of agent-specific scenarios, ranging from broad instruction alignment to highly specialized and efficient adaptation.

\textbf{RL} approaches refine LLM-based agents through interaction-driven optimization, where learning signals are derived from human feedback, learned reward models, or direct environmental responses, enabling adaptation beyond the supervised data regime. For instance, AgentGym \citep{agentgym} optimizes agent policies using objective functions defined over environment-generated trajectory rewards, whereas WebRL \citep{webrl} introduces a self-supervised Optimal Reward Model to assess trajectory quality and provide implicit evaluative signals for policy updates. Beyond such generic reward formulations, several studies design task-specific, multi-objective reward functions to guide agent behavior along additional dimensions, including computational or interaction efficiency \citep{agile}, policy stability through constrained updates \citep{copy}, and confidence calibration by penalizing overconfident erroneous outputs \citep{sayself}. Collectively, these RL-based methods demonstrate the adaptability of reward-centric optimization for deploying LLM-based agents in complex and evolving environments. Nevertheless, there remains no unified framework that effectively integrates these SFT, offline RL, and online RL paradigms to address multi-task agentic planning with strong robustness and generalization guarantees.

\subsection{Mixture-of-Experts (MoEs) Training and Stability}
As models scale up, Mixture-of-Experts (MoEs) architectures are increasingly favored for reducing inference costs and deployment overhead without sacrificing expressive power. However, the common ``token-choice" routing paradigm often leads to expert load imbalance, causing a few experts to process most tokens while others remain under-utilized.

Standard mitigation includes auxiliary load-balancing losses\citep{shazeer2017outrageously}, z-loss\citep{zoph2022st}, and expert capacity constraints \citep{lepikhin2020gshard, zoph2022st}. More recent innovations have introduced adaptive loss coefficients \citep{wei2024skywork}, auxiliary-loss-free load balancing \citep{wang2024auxiliary} and global-batch-level statistics \citep{qiu2025demons} to stabilize training. Some studies have focused on enhancing expert specialization by introducing orthogonal loss \citep{guo2025advancing} and ERC loss \citep{lv2025coupling}. Despite these advances, in the context of multi-domain agent training---where task formats range from task orchestration to long-horizon logic scheduling---naive balancing often suppresses expert specialization. Therefore, a training method that dynamically balances the specialization of heterogeneous task experts is essential for building generalize agent models based on MoE architectures.

\section{Data Generation}
The efficacy of agentic planning is fundamentally contingent upon the quality and diversity of the underlying training data. Although LLMs have demonstrated significant potential as autonomous agents, the scarcity of high-quality agent planning data remains a primary bottleneck to satisfy multi-type planning applications. Traditional data acquisition methodologies, which predominantly rely on labor-intensive human–agent interactions or computationally expensive sandbox-based simulations, suffer from limited scalability and often result in ``single-point optimizations" that fail to generalize diverse agentic tasks.

To address these challenges, we introduce a lightweight and scalable synthetic data generation framework, designed to significantly reduce the computational and manual overhead associated with synthesizing planning data. Our framework systematically encompasses the dominant paradigms of agentic planning: hierarchical task decomposition, tool-augmented planning, multi-constraint scheduling, procedural logic orchestration and long-horizon tool execution. The subsequent sections detail our methodology for synthesizing data across these varied scenarios, ensuring both the logical rigor and the operational stability necessary for high-performance agentic tuning.

\subsection{Hierarchical Task Decomposition} \label{HTD}

\begin{figure}
    \centering
    \includegraphics[width=0.9\linewidth]{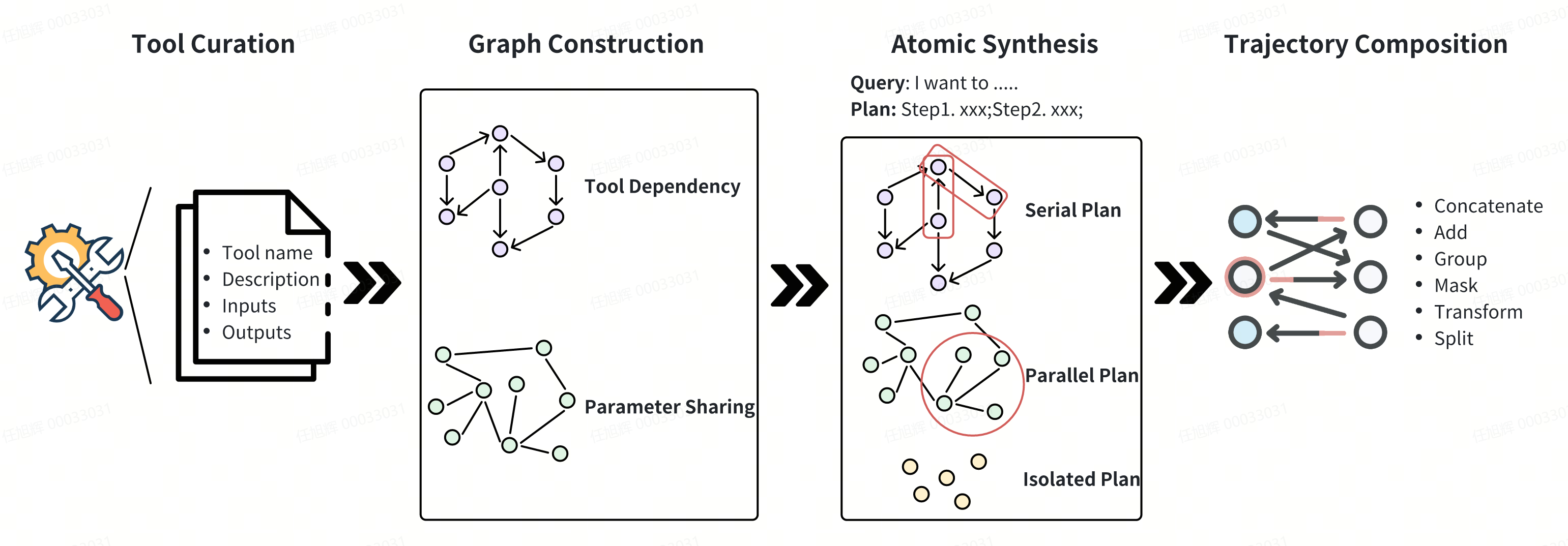}
    \caption{The synthesis pipeline of hierarchical task decomposition.}
    \label{fig:task_decomposition}
\end{figure}

The deployment of LLM agents in real-world environments is often impeded by the complexity of intricate user instructions. To address this, we employ hierarchical task decomposition as the primary structural mechanism to partition high-level user intents into granular, executable sub-tasks \citep{Magnet, prasad-etal-2024-adapt}. By decoupling complex planning into a structured hierarchy, this approach reduces the cognitive load on the executor, mitigates the risk of cascading error propagation, and significantly enhances the interpretability of the agent's execution trajectory.

To ensure that decomposed sub-tasks are both executable and sufficiently granular, we utilize a broad spectrum of functional APIs as the foundational building blocks of our planning formulation. Here, an executable ``sub-task" is rigorously defined as a target tool invocation. As illustrated in Figure~\ref{fig:task_decomposition}, our synthetic data pipeline for hierarchical task decomposition follows a four-stage process: (1) \textbf{Tool Curation}: Aggregating a high-fidelity repository of tools accompanied by exhaustive technical documentation. (2) \textbf{Graph Construction}: Modeling functional dependencies and relational synergies via tool relationship graphs. (3) \textbf{Atomic Synthesis}: Generating "Atomic Plans", the simplest units of execution derived from graph topology. (4) \textbf{Trajectory Composition}: Synthesizing complex, multi-turn dialogue trajectories by composing these atomic actions.

In the initial stage, we aggregated over 5,000 tools from open-source APIs and our internal repository, spanning diverse domains such as e-commerce, productivity (e.g., email), entertainment, and travel. To ensure semantic consistency across heterogeneous resources, we normalized all tool definitions to strictly adhere to the OpenAI tool schema.\footnote{https://platform.openai.com/docs/guides/function-calling} Furthermore, we refined the descriptions to incorporate explicit input-output specifications, ensuring each entry contains precise metadata: tool name, functional description, input parameters, and output schemas.

To capture the logical dependencies and co-occurrence probabilities between tools, we construct two distinct topological structures: a directed Tool Dependency Graph ($G_{depend}$) and an undirected Parameter Sharing Graph ($G_{share}$). $G_{depend}$ models the causal flow where the output of one tool is required as the input for another, while $G_{share}$ captures latent logical synergies and parameter overlap. We posit that conventional data generation methods, which often rely on stochastic random walks over tool graphs, fail to guarantee the semantic coherence of tool combinations. Consequently, we introduce the concept of the Atomic Plan, a minimal semantic unit consisting of one or two tool executions to control generation quality.

Conditioned on the topologies of $G_{depend}$ and $G_{share}$, we define three distinct strategies for synthesizing Atomic Plans, denoted as $U = \{q, p(q)\}$, where $q$ represents the query and $p(q)$ the corresponding plan:

\begin{itemize}

    \item \textbf{Isolated Plan ($U_{iso}$)}. This represents the simplest atomic unit containing a single sub-task.$$U_{iso} = \{q, \{f_1\}\}$$Here, the LLM generates a query based on a randomly sampled tool function $f_1$, utilizing its documentation and a set of few-shot prompts to ensure relevance.

    \item \textbf{Serial Plan ($U_{ser}$)}. This unit encapsulates two sub-tasks exhibiting a strict causal dependency.
    $$U_{ser} = \{q, \{f_1 \to f_2\}\}$$
    To generate a serial task, we sample a connected edge $(f_1, f_2)$ from the directed dependency graph $G_{depend}$. The documentation for both tools serves as the context for the LLM simulation, prompting the generation of a query where the execution of $f_1$ is a prerequisite for $f_2$.

    \item \textbf{Parallel Plan ($U_{par}$)}. This unit contains two sub-tasks that are logically related but executionally independent.
    $$U_{par} = \{q, \{f_1, f_2\}\}$$
    For this category, we sample two tools $f_1$ and $f_2$ from the graph that share a logical connection (e.g., via $G_{share}$) but lack a direct dependency edge in $G_{depend}$. The LLM generates a multi-objective query requiring the independent execution of both tools.

\end{itemize}

We posit that these atomic plans serve as the minimal semantic units for hierarchical task decomposition. To construct sophisticated, high-entropy planning scenarios that mirror real-world complexity, particularly for multi turn interactions, we introduce a set of operations to aggregate the atomic plans. These operators manipulate the topological structure of atomic plans to synthesize complex trajectories:

\begin{itemize}

    \item \textbf{Concatenate}.
    Constructs a linear dependency chain by sequentially linking atomic tasks that share a unified data flow. This operator enforces a strict input-output coupling, where the execution result of a preceding atomic unit serves as the requisite input for the subsequent unit. This process synthesizes long-horizon planning trajectories characterized by deep, transitive dependencies.

    \item \textbf{Add}.
    Aggregates semantically disjoint atomic plans into a single composite directive. By combining tasks that possess no mutual dependencies or shared parameters, this operator simulates complex ``multi-intent" user queries. This requires the agent to parse and execute multiple independent objectives simultaneously within a single conversational turn.
    
    \item \textbf{Group}.
    Identifies and fuses atomic tasks that operate on identical parameter spaces (as indicated by $G_{share}$). This operator consolidates redundant arguments, optimizing the context window and training the agent to recognize opportunities for parameter reuse and efficient context sharing across different sub-tasks.

    \item \textbf{Mask}.
    Selectively obscures obligatory parameters within a tool invocation schema to simulate information asymmetry. This operation forces the transition from direct execution to information-seeking behavior, compelling the agent to suspend execution and generate clarifying questions (slot-filling) to retrieve the missing constraints from the user.

    \item \textbf{Transform}.
    Converts the relationship between atomic plans into conditional edges ($f_1 \xrightarrow{c} f_2$). This introduces branching logic where the execution of subsequent tools is predicated on specific output states or boolean conditions returned by antecedent actions, thereby training the agent in dynamic, reactive planning.

    \item \textbf{Split}.
    Deconstructs a composite task description into a temporal sequence of interactions. Rather than presenting the full intent in the initial prompt, this operator distributes the constraints across multiple dialogue turns. This simulates a realistic user who iteratively refines their request or provides instructions in stages, requiring the agent to maintain state over long context windows.
    
\end{itemize}

Following the generation phase, we implement a rigorous quality assurance pipeline comprising three distinct filtering mechanisms: deduplication, schema validation, and semantic verification. First, we employ SimHash \cite{10141242572242592} to compute local sensitivity hashes, identifying and removing samples with high similarity to eliminate redundancy. Second, we enforce a strict structural schema for all entries:
\begin{center}
    \texttt{<Query>...</Query>}\texttt{\textbackslash n}\texttt{<Plan>...</Plan>}.
\end{center}
Any sample deviating from this format is automatically discarded. Finally, we utilize a state-of-the-art LLM (i.e., GPT-5.2) to detect potential semantic drift in the generated content. Only samples that satisfy all three validation criteria are retained for the final training corpus. The generated samples can refer to Appendix \ref{td_examples}.

\subsection{Tool-Augmented Planning}
While Large Language Models (LLMs) demonstrate robust logical reasoning capabilities, their efficacy in real-world applications is frequently constrained by the static nature of their parametric knowledge. Tool-augmented planning provides a critical mechanism to bridge this gap, enabling agents to extend their operational boundaries through the dynamic invocation of external interfaces and APIs. However, acquiring high-fidelity training data for such capabilities has historically presented a significant challenge, primarily due to the reliance on resource-intensive sandbox simulations and the prohibitive costs associated with human-in-the-loop annotation.

To circumvent these efficiency bottlenecks, we introduce a lightweight synthetic data generation framework tailored for tool-augmented scenarios, drawing upon the bidirectional translation methodology proposed in \cite{Magnet}. Leveraging the structured outputs from the Hierarchical Task Decomposition phase (described in Section~\ref{HTD}), we establish a streamlined pipeline: \textbf{Atomic Plan $\rightarrow$ Hierarchical Task Decomposition $\rightarrow$ Tool-Augmented Planning}. In this framework, we treat the hierarchical task decomposition data as the semantic specification—or "natural language blueprint" for the corresponding tool-call trajectories. This approach allows us to fully exploit the pre-defined topological relationships between tools (as modeled in $G_{depend}$ and $G_{share}$) to synthesize comprehensive interaction traces. Unlike stochastic methods, this manually defined and controllable task construction ensures that the resulting trajectory data exhibits superior stability, completeness, and adherence to logical constraints, effectively filtering out the noise inherent in simulation-based data collection.

To operationalize this blueprint, we instantiate each sub-task from the decomposition phase as a discrete, sequential tool invocation. However, while the decomposition data provides the ground-truth sequence of "golden" function calls ($a_t$), it lacks the intermediate cognitive context the reasoning traces ($e_t$) required for robust agent training. To reconstruct this missing modality, we adopt the \textit{ReAct} (Reasoning + Acting) paradigm \citep{react}. We employ a specialized Reasoning Agent to synthesize the thought process for each step. By conditioning this agent on the conversation history and the target golden tool call, we prompt it to articulate the logical rationale justifying the selection of $a_t$ given the current state. This "hindsight reasoning" ensures that the generated thoughts are perfectly aligned with the subsequent actions.

Simultaneously, generating valid environmental observations ($o_t$) presents a distinct challenge. Conventional approaches that rely on live API invocations are often plagued by latency, temporal instability, and access costs. To mitigate these issues, we draw inspiration from \citep{stableToolBench} and implement a virtualized API Simulator. This simulator mimics the behavior of real-world interfaces, providing deterministic and semantically consistent responses to tool calls. Consequently, our data generation pipeline functions as an end-to-end multi-agent system: the Reasoning Agent synthesizes the cognitive rationale, the Planner executes the golden actions, and the Simulator Agent generates the environmental feedback, yielding a complete $(e_t, a_t, o_t)$ trajectory for every turn.

To better emulate the complexity of real-world applications, our framework moves beyond linear task execution to support conditional planning. We define the topological relationship between atomic sub-tasks through two distinct connection types: static and dynamic. A static connection enforces a fixed execution order where task $T_{i+1}$ follows $T_i$ regardless of the intermediate output. Conversely, a dynamic connection introduces a branching logic, where the subsequent task or whether a task is executed at all is conditioned on the environmental feedback (observation) from the preceding step.

\begin{figure}[ht]
    \centering
    \includegraphics[width=\linewidth, keepaspectratio]{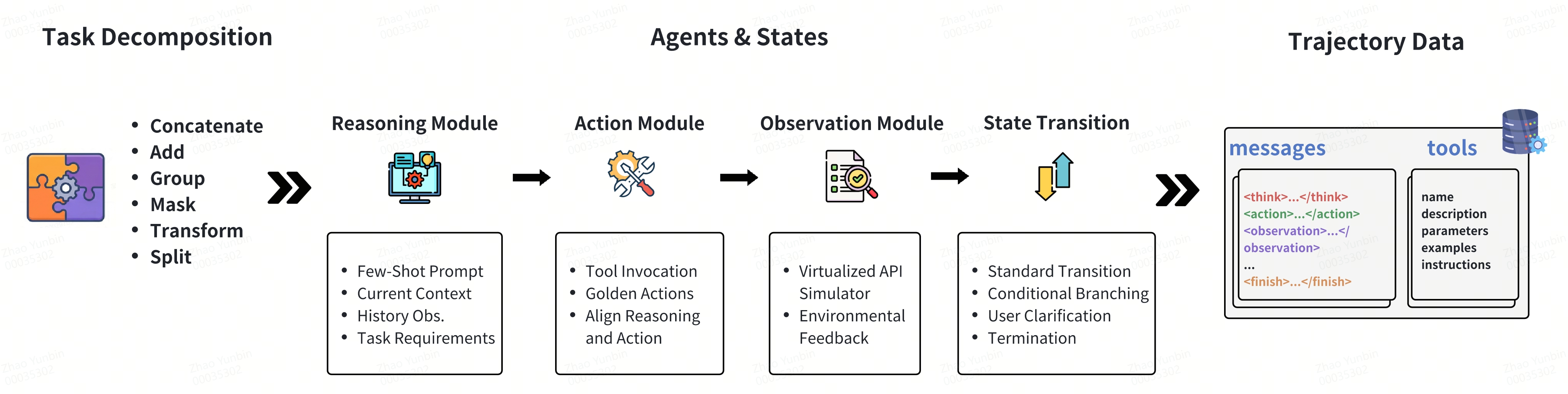}
    \caption{The synthesis pipeline of tool use planning.}
    \label{fig:tooluse_flowchart}
\end{figure}

We model the data generation process as a state transition system. For any given user query, the system synthesizes a trajectory composed of atomic turns. As illustrated in Figure~\ref{fig:tooluse_flowchart}, each turn is generated through a cycle of three specialized functional modules, producing a tuple $(e_t, a_t, o_t)$: \textbf{Reasoning Module} ($e_t$): employing a few-shot prompting strategy, this module synthesizes the "thought" process. It analyzes the current context, history, and task requirements to articulate the logical rationale for the next move, effectively bridging the gap between the user's intent and the tool's parameters. \textbf{Action Module} ($a_t$): This module determines the specific tool invocation. In our pipeline, this is derived from the "golden" actions identified in the hierarchical decomposition phase. We strictly validate consistency between the generated reasoning ($e_t$) and this ground-truth action to ensure alignment. \textbf{Observation Module} ($o_t$): To avoid the instability and latency of live API calls, we utilize the Virtualized API Simulator \citep{stableToolBench}. This module generates synthetic, semantically consistent environmental feedback based on the action, allowing for a robust and lightweight data synthesis pipeline without external network dependencies.

The progression of the trajectory is governed by a State Transition Mechanism, which determines the next system state based on the generated observation. We identify four transition categories:

\begin{itemize}
    \item \textbf{Standard Transition}: Proceed to the next sequential step in the static plan.

    \item \textbf{Conditional Branching}: Dynamically select the next sub-task based on the specific content of the observation.

    \item \textbf{User Clarification}: Trigger a request for user input if tool parameters are missing or ambiguous.

    \item \textbf{Termination}: Conclude the workflow and generate a final summary upon completing all required sub-tasks.
\end{itemize}

To ensure the high fidelity of the synthesized dataset, we implement a rigorous post-processing pipeline comprising syntactic verification, semantic auditing, and negative sampling. First, we enforce structural integrity by mandating valid JSON schemas and a strict thought-action-observation format: 
\begin{center}
    \texttt{<think>...</think>}, \texttt{<action>...</action>}, \texttt{<observation>...</observation>}
\end{center}
During this stage, null parameters are automatically pruned to minimize data noise. Second, an auxiliary LLM performs a semantic audit to validate causal consistency across the query, reasoning trace, and tool selection, discarding any trajectories that exhibit hallucinations or logical inconsistencies. Finally, to bolster model robustness, we employ negative sampling by injecting ``distractor tools" semantically proximate but functionally irrelevant metadata into the context. This forces the agent to differentiate between precise matches and plausible decoys. The format of the generated samples can refer to Appendix \ref{tc_examples}.

\subsection{Multi-Constraint Scheduling}
Multi-constraint scheduling entails the construction of detailed itineraries or resource allocations that optimize efficiency while adhering to a complex set of interacting constraints. This paradigm encapsulates critical real-world scenarios-such as travel logistics, meeting coordination, and calendar management-where the core challenge lies in simultaneously reasoning across diverse temporal, resource, and preference-based restrictions. Despite the potential of LLMs in automating administrative tasks, current models frequently falter when faced with the precise combinatorial reasoning required to satisfy interdependent constraints.

Given its inherent complexity, multi-constraint scheduling serves as a rigorous testbed for evaluating advanced reasoning capabilities. However, effective solutions for enhancing model performance in this domain remain scarce. Recent efforts have primarily focused on evaluation benchmarks. For instance, \citet{zheng2024natural} introduced NATURAL PLAN, a benchmark utilizing natural language tasks to evaluate the scheduling capacity of LLMs, observing significant performance gaps in state-of-the-art models. Similarly, \citet{xie2024travelplannerbenchmarkrealworldplanning} developed a sandbox environment to assess LLM performance in realistic travel planning scenarios.

\begin{figure}[ht]
    \centering
    \includegraphics[width=\linewidth, keepaspectratio]{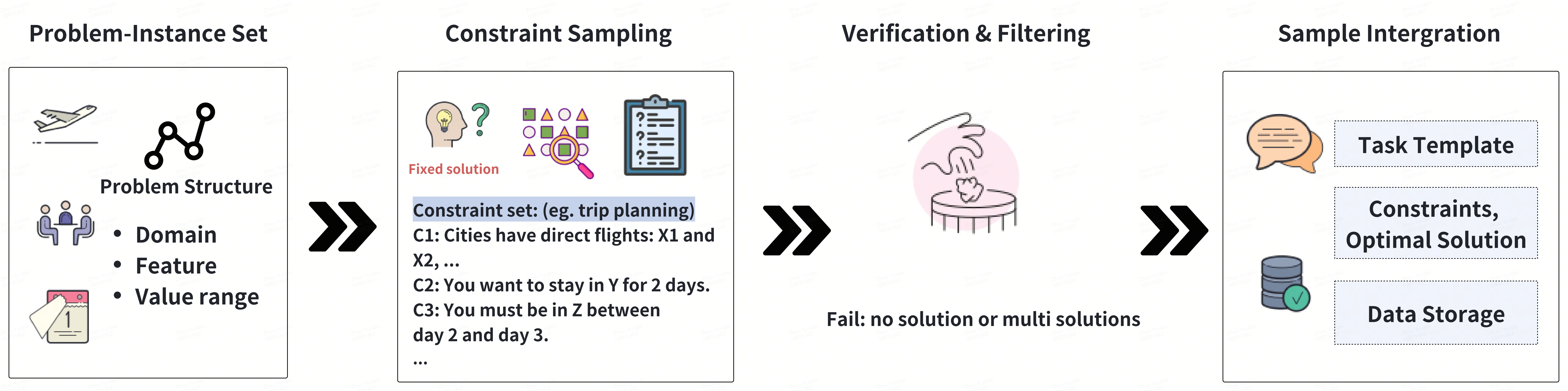}
    \caption{The synthesis pipeline of multi-constrain scheduling.}
    \label{fig:natural_plan_flowchart}
\end{figure}

Building upon these foundations, we formulate travel planning, meeting coordination, and calendar scheduling as instances of a unified Constraint Satisfaction Problem (CSP). To address the lack of high-quality training data, we propose a generalized framework designed to synthesize complex planning scenarios. The data generation workflow is illustrated in Figure~\ref{fig:natural_plan_flowchart} and consists of the following stages:

\begin{itemize}
    \item \textbf{Problem-Instance Construction.} We begin by initializing the problem space for each sample. Structural elements-such as the number of days, specific cities, or candidate flights—are instantiated by randomly drawing from a predefined entity database. These elements collectively define the boundaries of a concrete problem instance. 
    
    \item \textbf{Optional Goal-Solution Fixing.} To guarantee feasibility and uniqueness for complex tasks, we optionally employ an inverse generation strategy. In this step, we first generate a valid target solution and subsequently derive the necessary constraints to describe it. We apply this approach primarily to calendar scheduling, where deriving constraints from a fixed schedule ensures high-fidelity data construction.
    
    \item \textbf{Constraint Sampling.} Based on the initialized problem instance (and the fixed solution, where applicable), we sample specific constraints from a library of parameterized templates. These may include temporal restrictions (e.g., ``Stay in Hong Kong for 2 days"), resource availability (e.g., ``Direct flights exist between Beijing and Hong Kong"), or complex logic (e.g., ``Visit 4 distinct global cities over 21 days").
    \item \textbf{Verification and Filtering.} To ensure data quality, we implement a rigorous validation phase. We apply rejection sampling and automated logic checks to identify and discard inconsistent, ambiguous, or unsolvable constraint sets, retaining only those that yield a unique (or uniquely optimal) solution.
    \item \textbf{Sample Integration.} In the final stage, the verified constraints and the corresponding optimal solution are synthesized into a coherent natural language description using task-specific templates. The resulting sample, comprising the task description, user constraints, and the canonical solution, is then serialized for training or evaluation.
\end{itemize}

The final format of the generated samples can refer to Appendix \ref{mcs_examples}.

\subsection{Procedural Logic Orchestration}

Procedural Logic Orchestration involves organizing and coordinating tasks, resources, and schedules to achieve specific objectives efficiently\citep{agentwf}. Recently, there has been growing interest in leveraging large language models (LLMs) to support agentic workflows\citep{agentwf, knowagent, workflowllm}. However, complex orchestration scenarios often require rigorous structural dependencies that challenge the reasoning capabilities of general-purpose LLMs. \citep{qiao2025benchmarking} introduce a benchmark from graph workflow structures and find that the ability of LLM agents to predict graph plans is much worse than that of sequence plans. To address this, we focus on enhancing the fundamental ability of models to handle structured orchestration tasks.

\begin{figure}[ht]
    \centering
    \includegraphics[width=\linewidth, keepaspectratio]{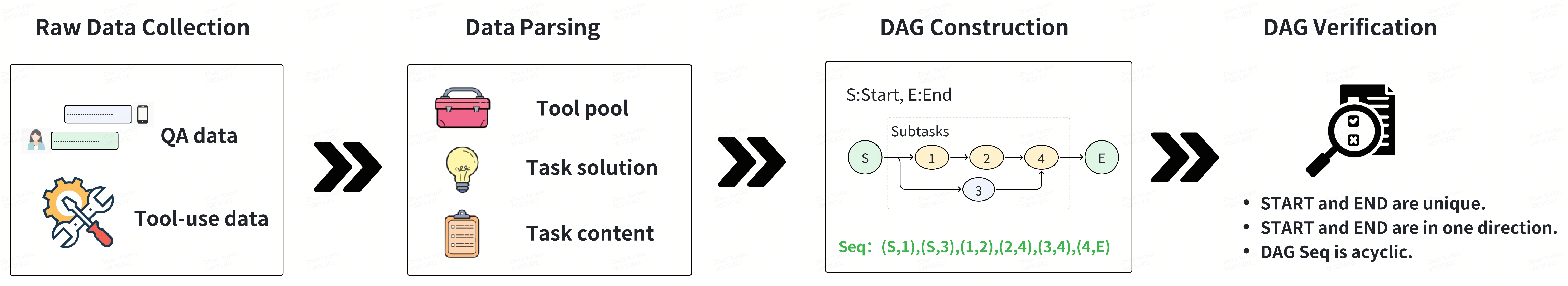}
    \caption{The synthesis pipeline of procedural logic orchestration.}
    \label{fig:workflow_planning_flowchart}
\end{figure}

In this section, we model workflow orchestration as a graph-orchestration problem, representing tasks as nodes and dependencies as directed edges in a directed acyclic graph (DAG). This formulation provides a robust testbed for evaluating and improving an agent's reasoning over complex dependencies. To advance capabilities in this domain, we propose a general framework for generating high-quality datasets for workflow orchestration. Our data-generation workflow comprises the following steps (also illustrated in Figure \ref{fig:workflow_planning_flowchart}):
\begin{itemize}
    \item \textbf{Raw Data Collection.} We collect two types of raw data: question-answering (QA) interactions and tool-use interactions. All raw data include gold-standard plans. Specifically, for QA we use the dataset released with \cite{li2025chainofagentsa}; for tool-use we use the dataset released with \cite{yin2024agenta}, augmented with our self-construct task-decomposition dataset. These sources form the basis for generating workflow orchestration instances.
    \item \textbf{Data Parsing.} We systematically parse the collected raw data to extract the elements essential for workflow orchestration. From each source we extract task descriptions and their corresponding solutions, consolidate all actions observed in the QA data into an action candidate pool, and aggregate the tools referenced in the tool-use data into a tool candidate pool.
    \item \textbf{DAG Construction.} For each sample, we retain actions or tools from the gold-standard plan and randomly select additional actions or tools from the candidate pools as distractors. We then construct a directed acyclic graph (DAG) representing the workflow plan based on the gold-standard plan, and derive a topological ordering of the DAG to obtain the final plan sequence.
    \item \textbf{DAG Verification.} To guaranty robust data quality, we apply stringent filtering criteria to discard invalid structures. First, we reject any DAG that fails to include all necessary actions or tools from the gold-standard plan. In addition, we filter out instances where the topologically derived sequence deviates from the ground truth. Finally, we eliminate any graphs containing cycles or failing to maintain unique START and END nodes, retaining only those candidates that satisfy all validity checks.
\end{itemize}

The format of generated data can refer to Appendix \ref{plo_examples}.

\subsection{Long-Horizon Tool Execution}
While the ability to invoke individual tools represents a foundational step in agentic intelligence, real-world autonomy frequently demands Long-Horizon Tool Execution—the capacity to orchestrate extended sequences of interdependent actions to achieve a complex objective. Unlike independent task decomposition or simple tool invocation, long-horizon scenarios introduce severe challenges regarding context retention and error propagation. In these settings, agents must not only select appropriate tools but also manage intricate tool chaining, where the output of one action serves as the prerequisite input for subsequent steps.

Current foundation models often falter in such environments due to limited operational memory and susceptibility to contextual drift, resulting in unstable execution trajectories. Some prior works propose to generate trajectory data based on manually crafted seed samples \cite{hu2025agen, Simia}. Although effective, they are labor intensive and time consuming for collecting and cleaning seed tasks. Inspired by these works and tackle the current limitations, we introduce a targeted trajectory generation framework designed to synthesize complex, multi-step tool interactions. This pipeline explicitly constructs scenarios requiring rigorous state tracking and logical dependency management. By procedurally generating data that enforces strict input-output coupling across extended temporal horizons, we aim to robustly enhance the model's execution stability and ability to maintain goal coherence throughout prolonged planning sessions.
 
\begin{figure}[ht]
    \centering
    \includegraphics[width=1.0\linewidth]{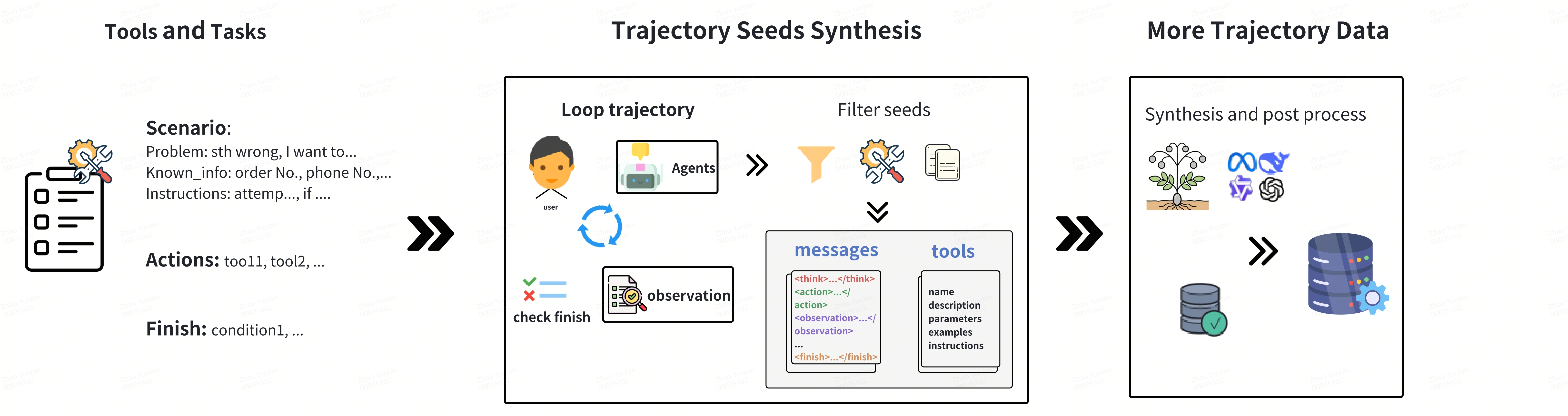}
    \caption{The synthesis pipeline of long-horizon tool execution.}
    \label{fig:long_horizon_flowchart}
\end{figure}

The synthesis pipeline consists of three stages: first, construct tasks with varying difficulty levels by appropriately combining atomic tasks. Next, generate long-context trajectory data involving multi-turn conversations and tool calls and filter high-quality samples as seed data. Finally, diffuse and generate more similar trajectory data. The details are in Figure~\ref{fig:long_horizon_flowchart}.

\textbf{Task Construction.}
To ensure that the constructed tasks are logical without contradictions or repetitions, the tasks are constructed based on atomic tasks with several steps. Firstly, atomic tasks are sampled according to the difficulty of the target task and the tools of them are collected in a list with the original relationships. Then, based on the selected atomic tasks, LLM is used to generate the initial setup $\mathit{Init}(requirement, known\_info, rules)$ for the task, including the user's problem and requirement, known information, the task rules that must be followed to solve the problem, and the conditions for completing the task. To ensure the long-horizon of the generated trajectory data, we set several requirements: 1. Number of tools must be greater than a threshold. 2. Initial user request: The request should not simply merge the queries of each atomic task, but a more general and vague requirement that requires reasoning to derive tool-related queries. 3. Termination conditions: All requests are fulfilled or more than 80\% of the candidate tools are used. Finally, manually verify the rationality and consistency of all tasks and requirements. 
In summary, a task consists of three parts: initial task setting, candidate tools, and terminal conditions.

\textbf{Long-Context Trajectory Seeds Synthesis.}
According to the task settings, we generate trajectory data several times for each task and filter high-quality samples as seed data. The synthesis pipeline is similar to the previous tool using planning data with some different parts:
\begin{itemize}
    \item The next query and task plan, $q_{t+1},T_{t+1}$ , are not fixed, but generated based on initial settings $\mathit{Init}$, historical queries, and historical trajectory data.
    \item Reasoning traces ($e_t$) and function calls ($a_t$) of each turn are generated by the Reasoning Agent and Action Agent according to the candidate tools.
    \item Data labeling: If the number of steps reaches the upper limit, or all tools are used but the problem is still not completely solved, the synthesis trajectory is labeled unsuccessful. If all requirements are met, it is labeled as successful.
\end{itemize}

The requirements for high-quality seed samples include: 1) fulfilling all task requirements, including the complete problem description and task specifications; 2) having the correct format, which follows a round structure of $\{ \text{query, (agent (thought,tool call), observation)} \times n, \text{agent (summary/reply)} \}$, and is saved in JSON format. To enhance the generalization ability of new samples, each seed sample further selects top task-related tools from the entire tools, together with the tools already used in the sample with each tool coming from a different atomic task.

\textbf{Scaling Data Generation}
To synthesize large scale data based on the seed samples, a prompt is organized with a random seed sample to generate the new sample with tasks and conversations similar. The system prompt consists of system role definition, core capabilities and execution logic, execution principles, candidate tools, sample trajectory data, and output format requirements. Each generation process completes all rounds of a sample, including user query, agent reasoning, tool-calling, environment feedback. Since the number of candidate tools is larger than that of the seed sample, the newly generated samples differ from the seed samples to some extent, ensuring diversity and robustness of the new samples. 

Finally, post-processing is also applied. The format of the new sample must be the same as that of the seed sample. Tool validation: the tool name and reasoning content are semantically consistent, the usage and definition of tool parameters are consistent, and the tool format follows the function format. The format of generated data can refer to Appendix \ref{lhte_examples}.

\section{Agent Foundation Model Training}
\begin{figure}
    \centering
    \includegraphics[width=1\linewidth]{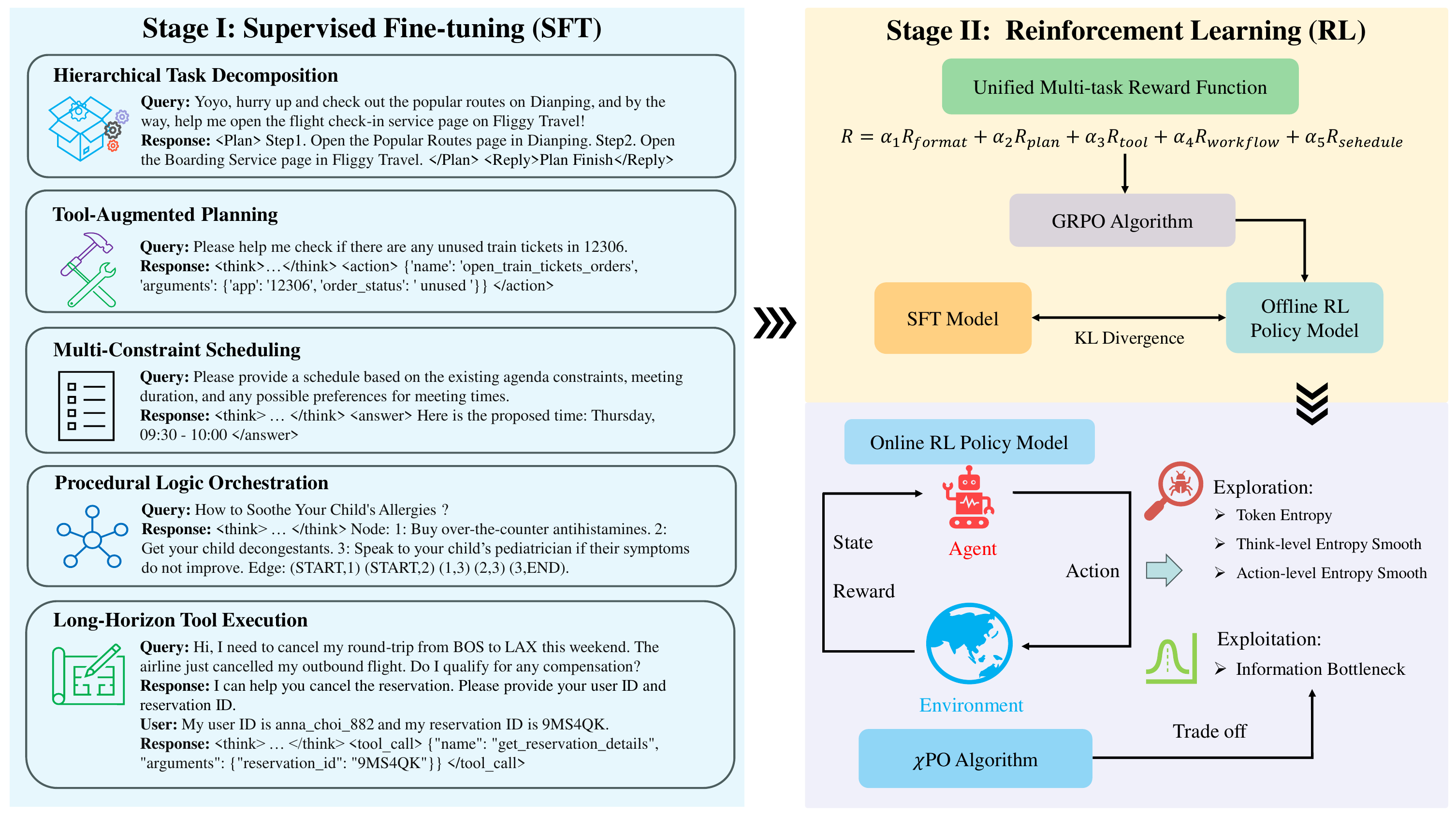}
    \caption{Overview of the MagicAgent training framework.}
    \label{fig:training_framework}
\end{figure}

Following the lightweight and scalable synthetic data generation framework, we have collected five types of planning data, including hierarchical task decomposition, tool-augmented planning, multi-constraint scheduling, procedural logic orchestration and long-horizon tool execution.  To achieve consistently remarkable performance across these heterogeneous tasks, we introduce MagicAgent, an end-to-end unified agent model designed to address hybrid planning tasks. The overall training framework is illustrated in Figure \ref{fig:training_framework} and consists of three stages: \textbf{Stage I: Supervised Fine-tuning (SFT)} (detailed in Section \ref{SFT}), which trains a foundation model on a large-scale and diverse planning dataset encompassing five fundamental planning tasks; \textbf{Stage II: Reinforcement Learning (RL)}, which further improves the accuracy and robustness of the model in heterogeneous static planning tasks (Section \ref{offline_RL}) and online environments under sparse task rewards (Section \ref{online_RL}).

\subsection{Agentic Supervised Fine-tuning}
\label{SFT}
We adopt the Qwen3 series as the base model and further train it using our synthetic dataset, including the Hierarchical Task Decomposition, Tool-Augmented Planning, Multi-Constraint Scheduling, Procedural Logic Orchestration, and Long-Horizon Tool Execution tasks. To ensure comprehensive coverage and diversity across domains, we employ the \emph{NovelSum} metric \citep{NovelSelect} to construct a representative sub-training set $\mathcal{D}_{sample}$, thereby facilitating the effective fusion of heterogeneous planning data. Specifically, the novelty of a candidate sample $x$ relative to the currently  selected training set $\mathcal{D}_{sample}$ is defined as follows:
\begin{align}
    v(x) = \sum_{x_{j} \in \mathcal{D}_{sample}} w(x, x_{j})^{\kappa_1} \cdot \sigma(x_{j})^{\kappa_2} \cdot d(x, x_{j}),
\end{align}
where the proximity weight $w(x, x_{j}) = 1/\pi(j)$ assigns higher importance to closer neighbors, with $\pi(j)$ indicating the rank of $x_{j}$ when samples are ordered by increasing distance to $x$. The density factor is defined as $\sigma(x_{j})=1/\sum_{k=1}^{K}d\left(x_j,N_k\left(x_j\right)\right) $, $N_k(x)$ denotes the $k$-th nearest neighbor of $x$, and $d(\cdot,\cdot)$ denotes the distance between sample embeddings. The hyper-parameters $\kappa_1$ and $\kappa_2$ control the degree of the proximity weight and density factor, respectively. At each iteration, the sample $x_{high}$ with the highest novelty score is selected with $x_{high}=\argmax_{x}v(x)$, and added to the training set as $\mathcal{D}_{sample} \leftarrow \mathcal{D}_{sample} \cup x_{high}$. This procedure is repeated, starting from $\mathcal{D}_{sample} = \varnothing$, until the predefined data budget is met.

The loss function formulation for the optimization objective of the SFT stage  is defined by as the standard token-level cross-entropy loss:
\begin{align}
    L_{\text{sft}} = -\frac{1}{N}\sum_{i=1}^{N}\sum_{j=1}^{V}y_{ij}log(p_{ij}),
    \label{sft_loss}
\end{align}      
where $N$ is the sequence length, $V$ is the vocabulary size, $y_{ij}$ is the ground-truth indicator for token $i$, and $p_{ij}$ is the predicted probability for class $j$ at token $i$. 
\subsection{Agentic Reinforcement Learning}
We define the agentic task as a finite-horizon Markov Decision Process (MDP) $M=\{\mathcal{T},S,A,P,R,\mu_0,H\}$, where
\begin{itemize} 
    \item{$\mathcal{T}$ denotes the type of tasks, including hierarchical task decomposition, tool-augmented planning, multi-constraint scheduling, procedural logic orchestration, long-horizon tool execution and online environment interaction.}
    \item{$S = (S^{data}, S^{env})$ represents the state space, comprising a static dataset component and an environment interaction component, where $S^{data}$ corresponds to user queries in the dataset and $S^{env}$ denotes the environment state.} 
    \item{$A$ is the vocabulary token space from which the agent generates plans or actions.} 
    \item{$P: S \times A \times S \rightarrow [0,1]$ defines the transition probability to the next state given the current state and action.} 
    \item{$R: S \times A \rightarrow \mathbb{R}$ is a unified, multi-task, rule-based reward function that measures the accuracy of the planning agent in completing heterogeneous tasks.} 
    \item{$\mu_0$ denotes the distribution over initial states, which may vary across tasks or environments.}
    \item{$H$ specifies the finite horizon of each task, where $H = 1$ corresponds to static tasks and $H > 1$ denotes the maximum number of interaction steps in the environment.}
\end{itemize} 
At each time step $t$, the planning agent observes a state $s_t \in S$, selects an action $a_t \in A$ according to a learned policy $\pi_{\text{RL}}(a_t|s_t;\theta_t)$, and then receives an immediate reward $r_t$.  The agent's objective is to learn an efficient policy that maximizes the expected return as $\mathbb{E}_{\pi_{RL}}(\sum_{t=0}^H r_t)$.

The primary challenge in training a model to perform diverse agentic tasks, including both static tasks and environment interactions, lies in the heterogeneity and interference among multiple tasks. To address this challenge, we propose a sequential reinforcement learning paradigm, as illustrated in  Figure~\ref{fig:training_framework}. Specifically, we combine static datasets and perform domain-wise reinforcement learning with verifiable rewards tailored to different tasks. The resulting static RL model is subsequently deployed and trained within an online robotic environment ALFWorld \citep{ALFWORLD}, enabling the model to generalize from offline reasoning tasks to embodied decision-making.

Furthermore, to better handle cross-task interactions in a multi-task setting, we design a unified multi-task verifiable reward function that evaluates both the output formats and the semantic correctness, thereby reducing inter-task interference. This unified reward function is described in detail in Section~\ref{reward_function}. In addition, we introduce heterogeneous, task-specific prompts to further distinguish tasks such as tool invocation, plan decomposition, and plan scheduling, which further improves task awareness and generalization. When the static training data corresponds to a specific task $\mathcal{T}_i \in \mathcal{T}$, we employ the standard Group Relative Policy Optimization (GRPO) algorithm to optimize the policy, as described in Section~\ref{offline_RL}. During the subsequent online training phase in the environment, we propose a novel reinforcement learning algorithm, termed $\chi$PO, which improves robustness and generalization in interactive settings by balancing exploration and exploitation. 
$\chi$PO incorporates additional entropy and information bottleneck regularization losses to stabilize training and prevent policy degeneration, as detailed in Section~\ref{online_RL}.

\subsubsection{Unified Multi-task Reward Function}
\label{reward_function}
Reinforcement Learning with Verifiable Rewards (RLVR) enables Large Language Models (LLMs) to generate responses and receive feedback from deterministic rule-based verifiers, as validated by DeepSeek-R1~\citep{deepseek_r1} and other models~\citep{magicgui,yue2025does,Tulu3}. In our agentic tasks, output format correctness, task planning, tool invocation, workflow construction, and schedule planning constitute five critical aspects for parsing valid planning actions. Accordingly, we design a unified multi-task, rule-based reward function composed of the following components:
\begin{align}
    R = \alpha_1 R_{format} + \alpha_2 R_{plan}  + \alpha_3 R_{tool} +  \alpha_4 R_{workflow} + \alpha_5 R_{schedule}.
\end{align}
Here, $\alpha_1, \alpha_2, \alpha_3, \alpha_4, \alpha_5 \in [0,1]$ denote the weighting coefficients of the respective reward components. For our five core planning tasks, the specific data format are summarized in Table~\ref{tab:reward}.

\begin{table}[!ht]
\centering
\caption{The unified multi-task data format.}
\vspace{.1cm}
\label{tab:reward}
\resizebox{\textwidth}{!}{
\small
\renewcommand{\arraystretch}{1.35}
\begin{tabular}{ll}
\toprule[1.5pt]
\textbf{Task} & \textbf{Data format} \\
\midrule[1pt]
Hierarchical Task Decomposition & $\texttt{<Plan>Step1.{\textcolor{blue}{P1}} Step2.{\textcolor{blue}{P2}}</Plan>}\texttt{<Reply>{\textcolor{blue}{R1}}</Reply>}$ \\
Tool-Augmented Planning & $\texttt{<think>...</think><action>{\textcolor{blue}{ToolList:[Tool1,...,Tooln]}}</action>}$ \\
Multi-Constraint Scheduling & $\texttt{<think>...</think><answer>{\textcolor{blue}{Schedule}}</answer>}$ \\
Procedural Logic Orchestration & $\texttt{Node 1:{\textcolor{blue}{N1}} 2:{\textcolor{blue}{N2}}. Edge:({\textcolor{blue}{E1}}),({\textcolor{blue}{E2}}),...,({\textcolor{blue}{En}})}$  \\
Long-Horizon Tool Execution & $\texttt{<think>...</think><tool\_call>{\textcolor{blue}{Tool}}</tool\_call>}$ \\
\bottomrule[1.5pt]
\end{tabular}}
\end{table}

\paragraph{Hierarchical Task Decomposition.}
The reward components are defined as follows:
\begin{flalign}
    & \quad\quad\quad\quad\quad R = \alpha_1 R_{format} + \alpha_2 R_{plan}, \nonumber \\
    & \quad\quad\quad\quad\quad R_{plan} = 1 \text{ if } \texttt{len(steps) == len(gt\_steps)} \text{ and } \texttt{cos\_sim(\{{P1},{P2},{R1}\}, gt)}  \ge \tau. & 
\end{flalign}
The format reward component $R_{format} = 1$ if the planning agent's output strictly adheres to the structure in Table ~\ref{tab:reward}. In addition, we employ a sentence embedding model to compute the cosine similarity between the generated content and the ground truth. Therefore, if the number of planning steps is correct and the cosine similarity scores of $\texttt{P1,P2,R1}$ with respect to the ground truth exceed predefined thresholds $\tau$, the planning reward component is set to $R_{plan} = 1$.

\paragraph{Tool-Augmented Planning.} The reward components for tool-augmented planning are defined as follows:
\begin{flalign}
        & \quad\quad\quad\quad\quad\quad\quad\quad\quad\quad\quad\quad\quad\quad\quad R = \alpha_1 R_{format} + \alpha_3 R_{tool}, \nonumber \\
        & \quad\quad\quad\quad\quad\quad\quad\quad\quad\quad\quad\quad\quad\quad\quad R_{tool} = 1 \text{ if } \texttt{{ToolList:[Tool1,...,Tooln]} == gt}. &
\end{flalign}
The format reward component is set to $R_{format} = 1$ if the output conforms to the above format. Tool-augmented planning data may invoke multiple tools within a single step. Therefore, the generated  \texttt{ToolList} must exactly match the ground truth in both tool identity and execution order, in which case the tool reward component is set to $R_{tool} = 1$.

\paragraph{Multi-Constraint Scheduling.} The reward components for multi-constraint scheduling are defined as follows:
\begin{flalign}
    & \quad\quad\quad\quad\quad\quad\quad\quad\quad\quad\quad\quad\quad\quad\quad R = \alpha_1 R_{format} + \alpha_5 R_{schedule}, \nonumber \\
    & \quad\quad\quad\quad\quad\quad\quad\quad\quad\quad\quad\quad\quad\quad\quad R_{schedule} = 1 \text{ if } \texttt{parse(schedule) == gt}. &
\end{flalign}
The format reward component is set to $R_{format} = 1$ if the planning agent's output satisfies the above structure. Furthermore, the generated $\texttt{Schedule}$, which may include travel logistics, meeting coordination, and daily calendar management, is parsed to extract key scheduling attributes such as time, location, and participants. If these attributes satisfy the specified constraints, the scheduling reward component is set to $R_{schedule} = 1$.

\paragraph{Procedural Logic Orchestration.} The reward components for procedural logic orchestration are defined as follows:
\begin{flalign}
    & \quad\quad\quad\quad\quad\quad\quad\quad\quad\quad\quad\quad\quad\quad\quad R = \alpha_1 R_{format} + \alpha_4 R_{workflow},  \nonumber \\
    & \quad\quad\quad\quad\quad\quad\quad\quad\quad\quad\quad\quad\quad\quad\quad R_{workflow} = \left(F_1\left(\texttt{node} \right) + F_1\left(\texttt{graph} \right) \right) / 2. &
\end{flalign}

The format reward component is set to $R_{format} = 1$ if the planning agent's output strictly conforms to the above structure. In addition, content evaluation is conducted under two complementary modes: node-level and graph-level. The node-level evaluation assesses the quality of the node sequence by emphasizing semantic similarity and relative ordering, which is quantified via the longest increasing subsequence (LIS) of matched nodes. The graph-level evaluation measures structural and semantic similarity by identifying the largest connected component formed by semantically matched nodes and shared edges. In both modes, a higher $F_1$-score indicates a higher-quality generated workflow.

\paragraph{Long-Horizon Tool Execution.} The reward components for tool execution are defined as follows:
\begin{flalign}
    & \quad\quad\quad\quad\quad\quad\quad\quad\quad\quad\quad\quad\quad\quad\quad R = \alpha_1 R_{format} + \alpha_3 R_{tool}, \nonumber \\
    & \quad\quad\quad\quad\quad\quad\quad\quad\quad\quad\quad\quad\quad\quad\quad R_{tool} = 1 \text{ if } \texttt{Tool == gt}. &
\end{flalign}
The format reward component is set to $R_{format} = 1$ if the output conforms to the above structure. For the tool-related component, tool invocations must strictly follow a valid JSON format, with all required brackets and quotation marks properly paired.Accordingly, the tool reward component is set to $R_{tool} = 1$ if the generated \texttt{Tool} exactly matches the ground truth. In addition, we introduce a special tool, $\texttt{reply("content"=...)}$, which is used to request missing information required for invoking other specific tools. For this reply tool, cosine similarity is also employed to evaluate the semantic accuracy of the generated content.

\subsubsection{Offline Reinforcement Learning}
\label{offline_RL}
For the static dataset, we employ the standard Group Relative Policy Optimization (GRPO) algorithm with a unified multi‑task reward function to optimize the policy $\pi_{\theta}(a \vert s)$, where $s$ denotes the state comprising both the system prompt and the user query, and $a$ represents either the generated plans or actions. The overall objective is defined as follows:
\begin{align}
    \mathcal{J}&_{\text{GRPO}}(\theta) = \mathbb{E}_{s \sim S, \{o_i\}_{i=1}^G\sim \pi_{\theta_\text{old}}(\cdot \vert s)} \nonumber \\
    & \Bigg[\frac{1}{G}\sum_{i=1}^{G}\frac{1}{|o_i|}\sum_{l=1}^{|o_i|} 
    \Bigg(\min \Big( \frac{\pi_{\theta}( o_{i,l} \vert s, o_{i, <l})}{\pi_{\theta_\text{old}}( o_{i,l} \vert s, o_{i, <l})} A_i, \text{clip} \big( \frac{\pi_{\theta}( o_{i,l} \vert s, o_{i, <l})}{\pi_{\theta_\text{old}}( o_{i,l} \vert s, o_{i, <l}) }, 1 - \varepsilon, 1 + \varepsilon \big) A_i \Big) - \beta D_{KL}(\pi_{\theta} \| \pi_{ref})\Bigg) \Bigg], \nonumber \label{J_GRPO}
\end{align}
where
\begin{align}
    A_{i} = \frac{r_i - {mean}(\{r_1,r_2,...,r_G\})}{{std}(\{r_1,r_2,...,r_G\})}, \quad D_{KL}(\pi_{\theta} \| \pi_{ref}) = \frac{\pi_{ref}( o_i \vert s)}{\pi_{\theta}( o_i \vert s)} - \log \frac{\pi_{ref}( o_i \vert s)}{\pi_{\theta}( o_i \vert s)} - 1.
\end{align}

Here, $mean$ and $std$ denote the mean and standard deviation of the group of rewards, respectively. The reward function used in this formulation is the Unified Multi-task Reward Function defined in Section \ref{reward_function}. Furthermore, the KL divergence term between the reference Supervised Fine-Tuning (SFT) policy $\pi_{ref}$ the current policy serves as a regularizer to promote stable learning in the offline reinforcement learning setting.

\subsubsection{Online Reinforcement Learning}
\label{online_RL}
In interactions with online environments, a primary challenge lies in balancing exploration and exploitation under sparse task reward conditions. Accordingly, we propose a novel \textbf{\emph{eXploration-eXploitation Policy Optimization}} ($\chi$PO) algorithm, which incorporates additional token-level entropy as well as think-level and action-level entropy smoothing regularization to regulate exploration. Simultaneously, an information bottleneck is introduced to enhance the accuracy and compactness of output actions, thereby promoting effective exploitation.

\paragraph{Exploration: Token-level Entropy Regularization.} In online reinforcement learning settings with sparse and delayed rewards, insufficient exploration often arises at the token generation level, where early overconfident predictions can prematurely collapse the policy into narrow response patterns. To explicitly encourage exploration, we introduce token-level entropy regularization, which injects controlled stochasticity directly into the language generation process. We view each generated token as a micro-decision that shapes the future interaction space. By maintaining higher entropy over the token distribution, the policy is encouraged to diversify lexical choices, reasoning trajectories, and intermediate representations, thereby expanding the reachable behavioral space prior to the observation of task-level rewards. Concretely, token-level entropy regularization is defined as follows:
\begin{align}
    \mathcal{J}^{\textit{token}}(\theta) &= \frac{1}{B}\sum\limits_{b=1}^{B} \frac{1}{H} \sum\limits_{h=1}^{H} \frac{1}{\vert o_{b,h} \vert} \sum\limits_{i=1}^{\vert o_{b,h} \vert} \mathcal{H}(o_{b,h,i}) 
    \label{token-entropy}
\end{align}
Here, $B$ is the batch size, $H$ represents the number of interaction steps in the environment, $\mathcal{H}(o_{b,h,i})$ is the entropy of the token $o_{b,h,i}$, and $\vert o_{b,h} \vert$ denotes the sequence length at step
$h$ of sample $b$. By embedding exploration directly into the token-generation mechanism, token-level entropy regularization provides a principled and scalable means of promoting exploration in online reinforcement learning, and serves as the foundation for higher-level entropy smoothing and information bottleneck strategies introduced later.

\paragraph{Exploration: Think-level and Action-level Entropy Smoothing Regularization.} While token-level entropy promotes local stochasticity, effective exploration in planning agents also depends on where uncertainty is permitted to accumulate. Inspired by the work EPO \citep{EPO}, we observe a pronounced scale mismatch between the average entropy of think segments and action segments: intermediate reasoning typically exhibits substantially higher entropy, whereas final actions naturally collapse toward low-entropy, decision-oriented outputs. Applying a uniform entropy constraint across both stages therefore results in suboptimal exploration dynamics.

To address this imbalance, we introduce a think-action disentangled entropy smoothing regularizer that explicitly assigns distinct exploration budgets to the two stages. For think-level tokens, a higher entropy ceiling is permitted, enabling the model to explore diverse reasoning trajectories, hypotheses, and intermediate abstractions, thereby promoting broader cognitive exploration prior to commitment. In contrast, action-level tokens are subject to a stricter entropy bound, preventing exploratory noise from propagating into final outputs where exploitation and decisiveness are paramount. Accordingly, the Entropy Smoothing Regularization is defined as follows:
\begin{align}
    & \quad \quad \quad \quad \quad \quad \mathcal{J}^{\textit{smooth}}(\theta) = \frac{1}{B}\sum\limits_{b=1}^{B} \frac{1}{H} \sum\limits_{h=1}^{H} \frac{1}{\vert o_{b,h} \vert} \sum\limits_{i=1}^{\vert o_{b,h} \vert} \mathcal{P}(o_{b,h,i}), \nonumber \\
    \mathcal{P}(o_{b,h,i}) & = 
        \begin{cases} 
            0, & \text{if } o_{b,h,i} \in \text{<think>...</think> and }   \alpha_{low}^{think}\bar{\mathcal{H}}^{think} \leq \mathcal{H}(o_{b,h,i}) \leq \alpha_{high}^{think}\bar{\mathcal{H}}^{think}, \\
            0, & \text{if } o_{b,h,i} \in \text{<action>...</action> and }   \alpha_{low}^{action}\bar{\mathcal{H}}^{action} \leq \mathcal{H}(o_{b,h,i}) \leq \alpha_{high}^{action}\bar{\mathcal{H}}^{action}, \\
            \rho, & \text{otherwise}.
        \end{cases}
    \label{smooth-entropy}
\end{align}
Here, $B$, $H$ and $\mathcal{H}(o_{b,h,i})$ follow the same notation as in Eq.(\ref{token-entropy}). $\bar{\mathcal{H}}^{think}$ and $\bar{\mathcal{H}}^{action}$ represent the  average historical entropy of think tokens and action tokens, respectively. Formally, the relation $\alpha_{low}^{think} < \alpha_{low}^{action} < 1 < \alpha_{high}^{action} < \alpha_{high}^{think}$ indicates that think-level smoothing tolerates and preserves higher entropy, thereby effectively extending the exploration horizon. $\rho$ is defined as a constant penalty term and is set to $\rho = -0.2$ in our experiments. Overall, think-action entropy smoothing transforms exploration from a flat, token-wise phenomenon into a hierarchical process, aligning exploratory effort with cognitive stages. This mechanism complements token-level entropy regularization by specifying where exploration should occur, thereby enhancing discovery under sparse rewards without compromising execution quality.

\paragraph{Exploitation: Information Bottleneck.} While the exploration strategy described above encourages the agent to discover diverse reasoning patterns, effective exploitation is equally critical for planning models to act reliably and efficiently. In this regard, we introduce an information bottleneck mechanism that explicitly favors action-relevant reasoning while suppressing redundant or non-informative intermediate thoughts. From an exploitation perspective, the goal is to retain only the minimal yet sufficient internal representations that support accurate decision-making during online reinforcement learning. Concretely, the information bottleneck regularizes the reasoning process by constraining the mutual information between the input prompt and the internal reasoning content, while simultaneously maximizing the mutual information between the reasoning and the resulting action. This trade-off encourages the agent to discard superfluous reasoning content that does not contribute to action selection, thereby improving exploitation efficiency. The corresponding regularization term is defined as follows:
\begin{align}
    \mathcal{J}^{\textit{IB}}(\theta) & = \mathcal{I}(\text{think}, \text{action})  - \gamma \mathcal{I}(\text{prompt}, \text{think})\nonumber \\
    & = \mathcal{H}(\text{action}) - \mathcal{H}(\text{action} \vert \text{think}) - \gamma \left[ \mathcal{H}(\text{think}) - \mathcal{H}(\text{think} \vert \text{prompt}) \right] \nonumber \\
    & \approx \gamma \mathcal{H}(\text{think} \vert \text{prompt}) - \mathcal{H}(\text{action} \vert \text{think}) + C.
    \label{information bottleneck}
\end{align}
Here, $\mathcal{I}(\cdot,\cdot)$ denotes the mutual information, $\gamma$ controls the exploitation-compression trade-off and is set to $\gamma = 1$ in our experiments, and $C = \mathcal{H}(\text{action}) - \gamma \mathcal{H}(\text{think})$ represents an approximately constant term. Maximizing this objective encourages the agent to compress its internal reasoning conditioned on the text input prompt while preserving the predictive capacity of the reasoning content for action generation. Consequently, the information bottleneck acts as a principled exploitation mechanism that aligns internal reasoning with actionable outcomes, resulting in more stable and effective planning behavior.

Finally, the overall objective of $\chi$PO is to maximize the following terms:
\begin{align}
    \mathcal{J}_{\chi\text{PO}}(\theta) = \mathcal{J}&_{\text{GRPO}}(\theta) + \lambda_1 \mathcal{J}^{\textit{token}}(\theta) + \lambda_2 \mathcal{J}^{\textit{smooth}}(\theta) + \lambda_3 \mathcal{J}^{\textit{IB}}(\theta),
\end{align}
where $\mathcal{J}_{\text{GRPO}}(\theta)$ follow the same notation as in Eq.(\ref{J_GRPO}), and $\lambda_1, \lambda_2, \lambda_3 \in (0,1)$ denote the weighting coefficients of the corresponding regularization terms.

\section{Load-Balanced Strategy for MoE Models}
To address the conflict between the robust planning capabilities of dense models and their prohibitive computational overhead for real-time deployment, we propose fine-tuning a Mixture-of-Experts (MoEs) variant of MagicAgent. By decoupling total parameter capacity from active FLOPs through sparse expert activation, the MoE architecture maintains high-dimensional representational power for complex tasks while reducing inference costs to levels comparable with smaller dense networks. This superior computation-to-parameter efficiency establishes MoE as a highly cost-effective solution for scalable agent deployment.

The core efficacy of an MoE model relies on its gating network (router), which dynamically dispatches input tokens to specialized experts. However, standard routing mechanisms often suffer from load imbalance, particularly in multi-task agent learning scenarios. Without regularization, routers tend to converge toward a "winner-takes-all" regime, where a small subset of experts processes the majority of tokens. This results in two failure modes: (1) Representation Collapse, where over-utilized experts generalized poorly across diverse tasks, and (2) Parameter Idleness, where under-utilized experts receive vanishing gradients, effectively reducing the model's capacity.

Addressing these challenges in the context of MagicAgent requires a tailored optimization strategy. Unlike general pre-training, our agent fine-tuning data consists of distinct, semantically homogeneous tasks (e.g., ierarchical task decomposition, tool-augmented planning, multi-constraint scheduling, procedural logic orchestration and long-horizon tool execution). We observe that applying standard sequence-level or micro-batch load balancing imposes overly rigid constraints. Since a single micro-batch often contains tokens from a single task, forcing the router to balance load locally disrupts expert specialization.

To mitigate these routing pathologies, we introduce a global load-balancing strategy combined with router stability regularization. This approach allows experts to specialize in distinct agent capabilities while maintaining computational efficiency across the global training distribution.

Formally, the total objective function for the MoE layer is defined as:
\begin{align}
{L}
= {L}_{\text{sft}}
+ \mu_1 \cdot {L}_{\text{gbl}}
+ \mu_2 \cdot {L}_z,
\end{align}
where ${L}_{\text{sft}}$ adopts the same supervised fine-tuning loss as in Eq. \ref{sft_loss}, ${L}_{\text{gbl}}$ represents the global batch load-balancing loss, and ${L}_{z}$ is the router z-loss. The scalars $\lambda_1$ and $\lambda_2$ are hyperparameters controlling the regularization strength.

Instead of enforcing balance within local micro-batches, we calculate expert usage statistics across the global batch (aggregated across all data parallel ranks). This global view encompasses a diverse mixture of tasks, allowing the model to balance expert utilization over the entire task distribution without penalizing task-specific specialization at the local level. The global load-balancing loss is formulated as:
\begin{align}
{L}_{\text{gbl}} ={N_E}\sum^{N_E}_{i=1}(\overline{f_i} \cdot \overline{P_i}),
\end{align}
where $N_E$ is the total number of experts, $\overline{f}_i$ represents the fraction of tokens dispatched to expert $i$ across the global batch, and $\overline{P}_i$ denotes the global average routing probability for expert $i$. This objective encourages the uniform distribution of tokens to experts over time, maximizing the effective parameter usage.

We further observed that during the early stages of mixed-task training, the router is prone to numerical overfitting on simpler tasks, producing disproportionately large logits. This logit instability suppresses exploration and hinders the learning of complex, long-horizon tool execution paths. To counteract this, we incorporate the z-loss regularization term to penalize large router logits and smooth gradient magnitudes:
\begin{align}
{L}_z
&= \frac{1}{T} \sum_{i=1}^{T}
\left(
\log \sum_{j=1}^{N_E} e^{m^{(i)}_j}
\right)^2,
\end{align}
where $T$ denotes the sequence length, and $m^{(i)}_j$ represents the logit assigned to the $j$-th expert for the $i$-th token. By constraining the magnitude of the logit matrix $\mathbf{M} \in \mathbb{R}^{T \times N_E}$, ${L}_z$ fosters robust routing decisions and ensures stable convergence across tasks of varying difficulty.

\section{Experiments}
To evaluate the planning capabilities of the MagicAgent series, we conduct a comprehensive comparative analysis against a representative set of industry-standard baselines, spanning both closed-source and open-source models. These models are stratified by parameter scale into two primary cohorts: (i) \textbf{Ultra-scale models} (exceeding 100B parameters), including GPT-5.2\citep{openai2025gpt52}, Kimi-K2-Instruct \citep{kimiteam2026kimik2openagentic}, DeepSeek-v3.1\citep{Dsv3}, GLM-4.7\citep{5team2025glm45agenticreasoningcoding}, Qwen3-235B-A22B-Instruct-2507\citep{Qwen3}, and Qwen3-Max \citep{qwen2025qwen3max}; and (ii) \textbf{Large-scale models} (within the 100B threshold), such as Qwen3-32B \citep{Qwen3}, Qwen3-30B-A3B-Instruct-2507\citep{Qwen3}, Llama3.3-70B-Instruct \citep{grattafior3herdmodels}, ERNIE-4.5-21B-A3B\citep{ernie2025technicalreport}, Olmo-3.1-32B-Instruct\citep{olmo2025olmo3}, and GLM-4.7-Flash\citep{zhipu2025glm47}. We first compare the performance on prominent open-source benchmarks in Section \ref{open-source-Benchmarks}. We further evaluate MagicAgent on our in-house MagicEval-Plan and MagicEval-Tool benchmarks, as presented in Section \ref{Magic-control}. Furthermore, we validate the online Reinforcement Learning (RL) $\chi$PO algorithm and the Mixture-of-Experts (MoEs) training paradigm and in Sections \ref{onlinerl-exp} and \ref{MOE-exp}, respectively. These experiments are designed to rigorously assess the planning effectiveness of our proposed MagicAgent models, with detailed training hyperparameters provided in Appendix \ref{training_details}.

\subsection{Performance of Open-source Benchmarks}
\label{open-source-Benchmarks}
\subsubsection{Benchmark Introduction}
The planning capabilities of the models are rigorously benchmarked using 11 specialized metrics across five key datasets: NaturalPlan \citep{zheng2024natural}, Worfbench \citep{qiao2025benchmarking}, $\tau^2$-Bench \citep{Tau2bench}, BFCL-v3 \citep{berkelenctioalliaderboard}, and ACEBench \cite{chen2025acebenchwinsmatchpoint}. These benchmarks are selected to provide an exhaustive evaluation of model performance in Tool-Augmented Planning, Multi-Constraint Scheduling, Procedural Logic Orchestration and Long-Horizon Tool Execution. The detailed illustration of each dataset is listed as follows:
\begin{itemize}
    \item \textbf{NaturalPlan (Multi-Constraint Scheduling):} This benchmark evaluates the model's ability to address complex scheduling tasks, emphasizing natural language planning scenarios such as trip planning and meeting scheduling, which require satisfying a dense set of spatial, temporal, and resource-related constraints. Evaluation on NaturalPlan assesses the model's deductive reasoning capabilities and its ability to maintain global consistency under complex scheduling requirements.

    \item \textbf{Worfbench (Procedural Logic Orchestration):} This benchmark is used to evaluate the model's proficiency in procedural logic orchestration. Unlike conventional reasoning tasks, Worfbench requires agents to navigate complex workflows involving conditional branching, iterative loops, and interdependent subtasks. It therefore provides a rigorous assessment of the model's ability to adhere to structured operational procedures and to manage control flow in multi-step processes, reflecting practical demands encountered in real-world planning and execution scenarios.

    \item \textbf{BFCL-v3 and ACEBench (Tool-Augmented Planning):} We employ the Berkeley Function Calling Leaderboard (BFCL-v3) and ACEBench to benchmark performance in tool-augmented planning. BFCL-v3 offers a robust framework for evaluating multi-tool function calling, requiring models to select appropriate APIs under dynamically evolving dialogue contexts. ACEBench complements this evaluation by introducing advanced tool-use scenarios that require higher-order planning to effectively bridge abstract user intents with concrete actions.

    \item \textbf{$\boldsymbol{\tau^2}$-Bench (Long-Horizon Tool Execution):} To evaluate the model's state-tracking and self-consistency capabilities, we employ $\tau^2$-Bench, a benchmark designed for long-horizon tool execution with realistic, multi-step tasks in domains such as retail and airline travel. It requires the agent to plan sequences of actions, adapt to dynamic feedback, and maintain execution fidelity over extended interaction cycles, thereby probing the limits of the model's operational memory, robustness, and long-term decision consistency in complex interactive environments.

\end{itemize}

To ensure fairness and reproducibility, we strictly followed the official instructions provided in the respective codebases for all baseline evaluations. Comprehensive details on the experimental setup are available in Appendix \ref{evaluationcode}.

\subsubsection{Main Results}

\begin{table}[!t]
\centering
\caption{Benchmark Performance Comparison Across Various Models. \textbf{Bold} indicates the best performance in each benchmark, while \underline{Underline} indicates the second-best performance.}
\vspace{.1cm}
\label{tab:model_comparison}
\resizebox{\textwidth}{!}{%
\begin{tabular}{lccccccccccc}
\toprule[1.5pt]
\multirow{2}{*}{\textbf{Models}} & \multicolumn{2}{c}{\textbf{WorfBench}} & \multicolumn{3}{c}{\textbf{NaturalPlan}} & \multicolumn{2}{c}{\textbf{$\boldsymbol{\tau^2}$-Bench}} & \multicolumn{2}{c}{\textbf{BFCL-v3}} & \multicolumn{2}{c}{\textbf{ACEBench}} \\
 & $F_1$ Chain & $F_1$ Graph & Trip & Meeting & Calendar & Retail & Airline & Live & Non-Live & En & Zh \\
\midrule[1pt]
\multicolumn{12}{l}{\textit{Ultra-Scale Models}} \\
\midrule[1pt]
GPT-5.2 & 60.1 & 39.1 & 39.4 & 53.7 & \textbf{64.5} & \textbf{75.2} & 50.5 & 76.8 & 82.2 & 70.7 & 80.1 \\
Kimi-K2-Instruct & 66.9 & 49.2 & 30.9 & 51.4 & 25.7 & 70.6 & 56.5 & 82.1 & 82.5 & 77.4 & 83.3 \\
GLM-4.7 & 67.3 & 49.8 & 31.3 & 42.7 & 46.4 & 66.0 & \textbf{64.5} & 80.8 & 86.6 & \underline{78.2} & \textbf{87.0} \\
DeepSeek-V3.1-nothink & 65.6 & 47.1 & 1.4 & 37.2 & \underline{64.4} & 72.8 & 51.5 & 71.1 & 67.6 & 67.8 & 73.2 \\
Qwen3-235B-A22B-Instruct & 63.2 & 44.7 & 21.9 & 31.7 & 53.5 & \underline{74.6} & 50.0 & 80.1 & 87.2 & 70.1 & 77.1 \\
Qwen3-MAX & 65.4 & 47.6 & 1.2 & 40.9 & 73.4 & 71.9 & \underline{57.5} & 82.5 & 89.1 & 77.6 & \underline{86.6} \\
\midrule[1pt]
\multicolumn{12}{l}{\textit{Large-Scale Models}} \\
\midrule[1pt]
Qwen3-32B-nothink & 65.8 & 49.2 & 37.6 & 21.5 & 31.9 & 48.8 & 24.0 & 80.1 & 86.3 & 57.2 & 70.1 \\
Qwen3-30B-A3B-Instruct & 62.4 & 44.4 & 19.4 & 20.3 & 38.4 & 57.0 & 38.0 & 75.8 & 85.3 & 61.3 & 71.1 \\
Llama3.3-70B-Instruct & 62.3 & 43.8 & 30.8 & 33.8 & 44.0 & - & - & 62.1 & 79.9 & 56.7 & 61.7 \\
ERNIE-4.5-21B-A3B-PT & 61.2 & 41.1 & 5.9 & 11.0 & 16.3 & 6.6 & 38.5 & 63.2 & 70.6 & 46.9 & 51.1 \\
Olmo-3.1-32B-Instruct & 33.2 & 22.7 & 14.1 & 20.2 & 26.2 & - & - & 44.4 & 42.4 & 25.1 & 18.4 \\
GLM-4.7-Flash & 64.5 & 47.4 & 5.9 & 16.9 & 21.3 & 54.2 & 55.0 & 77.3 & 82.9 & 53.5 & 62.0 \\
\midrule[1pt]
\multicolumn{12}{l}{\textit{MagicAgent Series}} \\
\midrule[1pt]
\rowcolor{blue!10} \textbf{MagicAgent-32B} & \textbf{80.3} & \underline{69.7} & \textbf{48.6} & \underline{57.7} & 61.5 & 62.0 & 53.0 & \textbf{84.1} & \textbf{89.7} & \textbf{78.3} & 84.1 \\
\rowcolor{blue!10} \textbf{MagicAgent-30B-A3B} & \underline{79.6} & \textbf{69.9} & \underline{42.8} & \textbf{60.7} & 58.3 & 53.1 & 55.0 & \underline{83.3} & \underline{89.6} & 74.9 & 83.5 \\
\bottomrule[1.5pt]
\end{tabular}%
}
\end{table}

The experimental results presented in Table \ref{tab:model_comparison} present a comprehensive evaluation of the MagicAgent series against state-of-the-art closed-source and open-source baselines. Our analysis focuses on four core capabilities: Multi-Constraint Scheduling, Procedural Logic Orchestration, Tool-Augmented Planning, and Long-Horizon Tool Execution. These capabilities are evaluated across five datasets: NaturalPlan, WorfBench, $\tau^2$-Bench, BFCL-v3, and ACEBench.

\paragraph{Procedural Logic Orchestration.} On the WorfBench benchmark, we evaluate performance using $F_1$ score at both the chain level and the graph level. The chain-level $F_1$ score assesses how well a model captures the longest increasing subsequence (LIS) of aligned nodes, jointly reflecting precision and recall while prioritizing semantic correspondence and correct relative ordering among the matched modes. In contrast, the graph-level $F_1$ score measures structural consistency by identifying the largest connected subgraph composed of semantically aligned nodes and their corresponding shared edges, thereby quantifying the quality of global relational alignment. The MagicAgent series exhibits a substantial performance improvement over both ultra-scale and large-scale models. In particular, MagicAgent-30B-A3B achieves the highest scores on both $F_1$ Chain ($79.6\%$) and $F_1$ Graph ($69.9\%$), surpassing GPT-5.2 by approximately $19.5\%$ and $30.8\%$, respectively. These results indicate that the proposed specialized training strategy enables more effective navigation of complex workflows, especially those involving conditional branching and interdependent subtasks, while maintaining stronger structural consistency than models with substantially larger parameters.

\paragraph{Multi-Constraint Scheduling.} On the NaturalPlan benchmark, which evaluates deductive reasoning and global consistency,  the MagicAgent series consistently achieves top-tier accuracy, with average scores of $55.9\%$ and $53.9\%$ for the Dense and MoE models, respectively. In the Trip and Meeting subset, MagicAgent-32B and MagicAgent-30B-A3B attain superior accuracy of $48.6\%$ and $60.7\%$, respectively, substantially outperforming GPT-5.2 ($39.4\%$ and $53.7\%$). In the Calendar subset, the MagicAgent models maintain consistently strong performance, effectively bridging the gap between large-scale efficiency and ultra-scale reasoning capabilities.

\paragraph{Tool-Augmented Planning.} The MagicAgent series demonstrates superior proficiency in tool-augmented planning, significantly outperforming ultra-scale closed-source models in both precision and multi-turn consistency.
\begin{itemize}
\item High-Precision API Invocation: On the BFCL-v3 benchmark, MagicAgent-32B achieves the highest reported Live Accuracy of $84.1\%$ and Non-Live Accuracy of $89.7\%$, with an average of $86.9\%$. This performance surpasses representative models such as Qwen3-Max ($82.5\%$) and GPT-4o ($66.9\%$), indicating a superior ability to map complex natural language intents to precise tool signatures.
\item Cross-Lingual Robustness: Results on ACEBench demonstrate substantial gains in both English (En) and Chinese (Zh) contexts. Notably, MagicAgent-32B achieves an average accuracy of $81.2\%$, exceeding DeepSeek-V3.1 ($67.8\%$ in En, $73.2\%$ in Zh) and Qwen3-235B ($70.1\%$ in En, $77.1\%$ in Zh) despite a significant disparity in parameter scale. These results highlight the effectiveness of planning-specific fine-tuning in distilling high-order reasoning into compact architectures.
\end{itemize}

\paragraph{Long-Horizon Tool Execution.} Evaluation via $\tau^2$-Bench assesses models' state-tracking and self-consistency capabilities over extended interaction horizons in the Retail and Airline domains. MagicAgent-32B and MagicAgent-30B-A3B achieve average accuracies of $57.5\%$ and $54.1\%$, respectively.

\begin{itemize}
    \item Adaptive State Tracking: In the Retail domain, MagicAgent-32B achieves an accuracy of $62.0\%$, corresponding to a $27.0\%$ relative improvement over the base Qwen3-32B model ($48.8\%$). This indicates a more robust internal representation of task state, effectively mitigating error accumulation over long horizons.
    \item Competitive Operational Integrity: Within the Airline domain, MagicAgent-30B-A3B reaches an accuracy of $55.0\%$,  exceeding substantially larger models such as Qwen3-235B-A22B ($50.0\%$) and closely approaching GLM-4.7-Flash. Its capacity to handle dynamic feedback and multi-step execution logic supports its suitability as a foundation for autonomous planning agents.
\end{itemize}

\begin{table}[!t]
\centering
\caption{MagicEval-Plan Performance Comparison Across Various Models. \textbf{Bold} indicates the best performance in each benchmark, while \underline{Underline} indicates the second-best performance.}
\vspace{.1cm}
\label{tab:Magic_plan}
\resizebox{\textwidth}{!}{%
\begin{tabular}{lcccccccccccc}
\toprule[1.5pt]
\multirow{2}{*}{\textbf{Models}} & \multicolumn{3}{c}{\textbf{General}} & \multicolumn{3}{c}{\textbf{Dependency}} & \multicolumn{3}{c}{\textbf{Condition}} & \multicolumn{3}{c}{\textbf{Context Inheritance}}  \\
 & Step & Embedding & LLM & Step & Embedding & LLM & Step & Embedding & LLM & Step & Embedding & LLM  \\
\midrule[1pt]
\multicolumn{13}{l}{\textit{Ultra-Scale Models}} \\
\midrule[1pt]
GPT-5.2 & 36.8 & 89.7 & \textbf{92.0} & 28.5 & 89.9 & \textbf{96.5} & 51.0 & 89.9 & 85.3 & 33.5 & 89.7 & \underline{91.8} \\
Kimi-K2-Instruct & 85.7 & 91.9 & 86.7 & 79.5 & 91.7 & 88.1 & 87.0 & 91.0 & 80.2 & 88.7 & 91.7 & 88.3 \\
GLM-4.7 & 87.9 & \textbf{97.0} & 88.1 & 82.5 & \underline{96.8} & 89.1 & 85.5 & \textbf{96.6} & 83.2 & \underline{94.6} & \textbf{96.6} & 88.2 \\
DeepSeek-V3.1-nothink  & 81.0 & \underline{96.8} & 84.0 & 64.0 & \textbf{96.9} & 88.3 & 65.0 & \underline{96.5} & 77.0 & 89.5 & \underline{96.3} & 89.9 \\
Qwen3-235B-A22B-Instruct & 70.6 & 91.4 & 85.1 & 65.5 & 90.5 & 87.0 & 72.5 & 89.6 & 74.6 & 82.1 & 90.8 & 88.0 \\
Qwen3-MAX & 81.0 & 92.7 & 89.3 & 78.0 & 91.8 & 88.8 & 76.0 & 91.0 & 80.6 & 88.3 & 92.0 & 90.7 \\
\midrule[1pt]
\multicolumn{13}{l}{\textit{Large-Scale Models}} \\
\midrule[1pt]
Qwen3-32B-nothink & 64.5 & 91.6 & 87.1 & 72.5 & 91.6 & 88.6 & 67.5 & 90.4 & 77.1 & 67.7 & 90.7 & 87.5 \\
Qwen3-30B-A3B-Instruct & 35.1 & 90.3 & 78.7 & 16.0 & 91.5 & 91.8 & 46.5 & 89.8 & 74.4 & 33.5 & 90.3 & 78.7 \\
Llama3.3-70B-Instruct & 36.8 & 92.1 & 82.6 & 38.5 & 90.7 & 86.4 & 40.5 & 87.9 & 70.2 & 38.5 & 90.3 & 85.5  \\
ERNIE-4.5-21B-A3B-PT & 72.7 & 92.4 & 85.1 & 71.0 & 91.6 & 86.6 & 49.0 & 90.2 & 76.2 & 76.3 & 91.3 &  87.9 \\
Olmo-3.1-32B-Instruct & 48.5 & 91.5 & 82.4 & 47.5 & 90.5 & 89.5 & 39.5 & 87.7 & 71.9 & 61.9 & 90.5 & 79.9 \\
GLM-4.7-Flash & 48.5 & 86.8 & 72.6 & 46.5 & 87.1 & 79.0 & 37.0 & 83.8 & 65.6 & 56.4 & 85.9 & 73.1  \\
\midrule[1pt]
\multicolumn{13}{l}{\textit{MagicAgent Series}} \\
\midrule[1pt]
\rowcolor{blue!10} \textbf{MagicAgent-32B} & \textbf{98.0} & 95.6 & \underline{91.2} & \textbf{98.2} & 95.7 & 91.2 & \textbf{97.5} & 95.3 & \underline{87.4} & \textbf{97.6} & 94.3 & \textbf{92.2} \\
\rowcolor{blue!10} \textbf{MagicAgent-30B-A3B} & \underline{97.6} & 95.9 & 91.0 & \underline{97.0} & 95.6 & \underline{92.3} & \underline{97.0} & 95.5 & \textbf{88.1} & \textbf{97.6} & 94.6 & 90.2 \\
\bottomrule[1.5pt]
\end{tabular}%
}
\end{table}

\subsection{Performance of in-house Benchmarks}
\label{Magic-control}
\subsubsection{Benchmark Introduction}
To further evaluate the real-world applicability of our model, we collected actual user interaction logs from our production environment to construct an in-house evaluation dataset, named MagicEval\footnote{All interaction data used for evaluation has been fully anonymized, with all personally identifiable information (PII) removed.}. Within our deployment scenario, the agent is expected to execute two primary types of tasks: user query decomposition and tool invocation planning. These tasks correspond directly to evaluating the agent's proficiency in hierarchical task decomposition and tool-augmented planning, respectively.

The evaluation covers four dominant conversational scenarios: (1) \textbf{General}: Queries are randomly sampled from the entire interaction corpus to assess the model's overall baseline performance across diverse conditions. (2) \textbf{Dependency}: Queries require the execution of sequential actions, specifically designed to test the model's ability to handle complex task dependencies. (3) \textbf{Condition}: Queries involve conditional execution paths, evaluating the model's capacity for dynamic decision-making based on specific constraints. (4) \textbf{Context Inheritance}: Queries necessitate the tracking and inheritance of information across multi-turn interactions, testing the model's comprehension of historical context.

\begin{itemize}
    \item \textbf{MagicEval-Plan (Hierarchical Task Decomposition):} To assess the efficacy of hierarchical planning, the MagicEval-Plan dataset evaluates the model's capacity to decompose high-level user queries into sequences of executable sub-tasks. Performance is quantified using a multi-dimensional approach: Step Accuracy validates the structural alignment of the generated plan length; Semantic Similarity utilizes embedding distances to ensure the generated tasks semantically match the annotated ground truth; and LLM-as-a-Judge leverages a state-of-the-art model (GPT-5.2) to provide a qualitative assessment of plan correctness. Collectively, these metrics serve to verify that the generated decomposition faithfully reflects human-annotated intent in terms of both structure and semantics.
    
    \item \textbf{MagicEval-Tool (Tool-Augmented Planning):} This dataset focuses on the precision of execution planning, specifically examining the model's ability to invoke internal tools correctly within a dialogue context. This evaluation relies on two industry-standard metrics: Tool Name Accuracy, which verifies that the correct function is selected from the available library, and Tool Argument Accuracy, which ensures the parameters passed to the function align perfectly with the required schema. These metrics are critical for benchmarking the reliability of the agent in production environments where precise and error-free API execution is paramount.
\end{itemize}

\subsubsection{Main Results}
\begin{table}[!t]
\centering
\caption{MagicEval-Tool Performance Comparison Across Various Models. \textbf{Bold} indicates the best performance in each benchmark, while \underline{Underline} indicates the second-best performance.}
\vspace{.1cm}
\label{tab:Magic_tool}
\resizebox{0.9\textwidth}{!}{%
\begin{tabular}{lcccccccc}
\toprule[1.5pt]
\multirow{2}{*}{\textbf{Models}} & \multicolumn{2}{c}{\textbf{General}} & \multicolumn{2}{c}{\textbf{Dependency}} & \multicolumn{2}{c}{\textbf{Condition}} & \multicolumn{2}{c}{\textbf{Context Inheritance}}  \\
 & Name & Argument & Name & Argument & Name & Argument & Name & Argument  \\
\midrule[1pt]
\multicolumn{9}{l}{\textit{Ultra-Scale Models}} \\
\midrule[1pt]
GPT-5.2 & 74.6 & 55.8 & 67.8 & 46.5 & 73.5 & 58.2 & 71.0 & 55.5  \\
Kimi-K2-Instruct & 89.9 & 66.8 & 91.0 & 58.3 & 88.7 & 71.0 & 90.3 & 67.5 \\
GLM-4.7 & 89.1 & 70.6 & 85.3 & 56.5 & 82.6 & 65.6 & 85.5 & 66.0  \\
DeepSeek-V3.1-nothink  & 86.8 & 65.0 & 86.8 & 54.8 & 86.6 & 66.5 & 86.3 &  62.0 \\
Qwen3-235B-A22B-Instruct & 89.3 & 67.3 & 89.5 & 55.3 & 87.5 & 69.2 & 88.8 &  63.8  \\
Qwen3-MAX & 92.4 & 72.3 & 90.0 & 58.3 & 88.1 & 69.5 & 89.5 & 67.0  \\
\midrule[1pt]
\multicolumn{9}{l}{\textit{Large-Scale Models}} \\
\midrule[1pt]
Qwen3-32B-nothink & 69.5 & 50.7 & 69.8 & 43.8 & 57.6 & 44.2 & 70.0 &  48.0 \\
Qwen3-30B-A3B-Instruct & 48.3 & 34.5 & 49.3 & 30.0 & 40.2 & 29.7 & 46.6 & 29.2 \\
Llama3.3-70B-Instruct & 74.9 & 53.8 & 71.0 & 41.8 & 76.8 & 53.1 & 75.5 &  48.8 \\
ERNIE-4.5-21B-A3B-PT & 71.6 & 43.7 & 61.5 & 32.8 & 61.3 & 40.2 &  74.5 & 43.3  \\
Olmo-3.1-32B-Instruct & 19.5 & 11.7 & 21.3 & 9.5 & 10.7 & 5.2 & 2.6 & 14.0  \\
GLM-4.7-Flash & 38.3 & 26.4 & 24.8 & 13.0 & 27.4 & 20.1 & 34.5 & 21.5 \\
\midrule[1pt]
\multicolumn{9}{l}{\textit{MagicAgent Series}} \\
\midrule[1pt]
\rowcolor{blue!10} \textbf{MagicAgent-32B} & \underline{97.5} & \textbf{87.3} & \underline{97.5} & \textbf{80.0} & \textbf{95.4} & \textbf{88.7} & \textbf{98.8} & \textbf{85.5} \\
\rowcolor{blue!10} \textbf{MagicAgent-30B-A3B} & \textbf{97.7} & \underline{86.3} & \textbf{98.5} & \underline{78.5} & \underline{94.2} & \underline{87.2} & \underline{98.5} & \underline{82.0} \\
\bottomrule[1.5pt]
\end{tabular}%
}
\end{table}

\paragraph{Hierarchical Task Decomposition.} Table \ref{tab:Magic_plan} evaluates model's hierarchical task decomposition ability under four structured planning settings. The results demonstrate that the MagicAgent series substantially improves fine-grained plan construction quality across all evaluation protocols (Step, Embedding, and LLM).

\begin{itemize}
    \item Step-based evaluation: Under the Step-level metric, which directly measures structural correctness of decomposed plans, MagicAgent achieves near-perfect performance across all four scenarios. MagicAgent-32B obtains $98.0\%$, $98.2\%$, $97.5\%$, and $97.6\%$, while MagicAgent-30B-A3B consistently reaches $97.0$. In contrast, even ultra-scale models exhibit noticeable structural weaknesses, particularly under Dependency and Condition settings where step accuracy often drops to $70\% \sim 90\%$. This gap indicates that large general-purpose models struggle with precise hierarchical decomposition when multi-step constraints are involved.
    \item Embedding-based evaluation: MagicAgent maintains stable and leading performance ($95\%$ across all tasks), suggesting strong semantic alignment between generated plans and reference decompositions. Importantly, unlike some ultra-scale models that show relatively high embedding similarity but lower step correctness, MagicAgent achieves both structural validity and semantic consistency.
    \item LLM-based evaluation: MagicAgent remains competitive across all scenarios, achieving balanced performance without sacrificing structural rigor. While certain ultra-scale models occasionally obtain slightly higher LLM scores on isolated tasks, they fail to match MagicAgent's consistent dominance on step-level correctness, which is more directly aligned with hierarchical decomposition fidelity.
\end{itemize}

Overall, the results indicate that MagicAgent substantially strengthens hierarchical task decomposition, producing plans that are not only semantically plausible but structurally precise. The consistent improvements across General, Dependency, Condition, and Context Inheritance settings suggest robustness under complex hierarchical constraints.

\paragraph{Tool-Augmented Planning.} Table \ref{tab:Magic_tool}  evaluates tool-augmented planning ability, decomposed into tool name and argument accuracy, across the same four structured reasoning scenarios. The results show that MagicAgent significantly improves both tool identification accuracy and parameter precision.

\begin{itemize}
    \item Tool name: MagicAgent achieves near-ceiling performance across all categories. Both variants consistently reach $95\% \sim 99\%$, clearly outperforming ultra-scale baselines whose best accuracy remain around $90\% \sim 92\%$. The advantage becomes more evident under Dependency and Context Inheritance settings, where correct tool selection requires reasoning over cross-step constraints. This indicates that MagicAgent more reliably integrates structural plan context into tool decisions.
    \item Tool Argument: For argument generation, which requires precise semantic grounding and constraint satisfaction, the performance gap becomes even more pronounced. MagicAgent achieves $80\% \sim 89\%$ across all settings, while most ultra-scale models remain below $70\%$. The largest improvements appear in the Dependency setting, suggesting that MagicAgent better propagates intermediate constraints into tool parameters, reducing argument-level inconsistencies.
\end{itemize}
Taken together, the results demonstrate that MagicAgent enhances both macro-level tool selection and micro-level argument specification. Unlike ultra-large models that may select plausible tools but produce partially incorrect arguments, MagicAgent shows stronger alignment between plan structure and tool execution details. This confirms its effectiveness in tool-augmented planning under structured, multi-constraint scenarios.

\definecolor{grayrlvmr}{RGB}{189,189,189}
\subsection{Performance of Online RL}
\label{onlinerl-exp}

\begin{table}[!t]
    \centering
    \caption{Performance comparison in the online ALFWorld environment. We report two success metrics: $\textbf{Succ.}^*$ and $\overline{\textbf{Succ.}}$. The table is organized into two parts: (i) a ReAct-based comparison across ultra-scale, large-scale, and MagicAgent series, and (ii) a various training-algorithm evaluation on the Qwen2.5 backbone. \textbf{Bold} and \underline{underline} indicate the best and second-best results, respectively, applied separately within these two parts. Results for baseline methods (ReAct, Agentgym, SFT, GRPO, GiGPO, RLVMR, EPO) are taken from the EPO paper. The \textcolor{grayrlvmr}{gray} color indicates that GIGPO and RLVMR employ the Qwen2.5-7B model, whereas EPO and our proposed $\chi$PO use the smaller-parameter Qwen2.5-3B model while achieving comparable performance. We run our own implementations of $\chi$PO across multiple random seeds to ensure stable results.}
    \vspace{.1cm}
    \label{tab:onlinerl_result}
    \small
    \begin{tabular}{lccccc}
    \toprule[1.5pt]
    \textbf{Models} & \textbf{Methods} & \textbf{IID} ${\textbf{Succ.}}^*$ & \textbf{IID} $\overline{\textbf{Succ.}}$ & \textbf{OOD} ${\textbf{Succ.}}^*$ & \textbf{OOD} $\overline{\textbf{Succ.}}$  \\
    \midrule[1pt]
    \multicolumn{6}{l}{\textit{Ultra-Scale Models}} \\
    \midrule[1pt]
    GPT-4o & ReAct & 57.3 & - & 66.0 & - \\ 
    GPT-5.2 & ReAct & 63.5 & - & 62.6 & - \\ 
    DeepSeek-V3.1 & ReAct & 67.1 & - & \underline{76.1} & - \\
    DeepSeek-R1 & ReAct & \underline{68.8} & - & 70.2 & - \\ 
    Qwen3-235B-A22B-Instruct & ReAct & 57.1	& - & 55.9 & - \\
    Qwen3-MAX & ReAct & 50.7 & - & 59.7 & - \\
    \midrule[1pt]
    \multicolumn{6}{l}{\textit{Large-Scale Models}} \\
    \midrule[1pt]
    Qwen3-32B & ReAct & 61.4 & - & 68.7 & - \\
    Qwen3-30B-A3B & ReAct & 44.3 & - & 44.0 & - \\
    Qwen3-8B & ReAct & 51.4 & - & 53.7 & - \\
    Qwen2.5-7B & ReAct & 23.1 & - & 28.5 & -\\ 
    \midrule[1pt]
    \multicolumn{6}{l}{\textit{MagicAgent Series}} \\
    \midrule[1pt]
    \rowcolor{blue!10} \textbf{MagicAgent-32B} & ReAct & \textbf{81.4} & - & \textbf{78.4} & - \\
    \rowcolor{blue!10} \textbf{MagicAgent-30B-A3B} & ReAct & 67.9 & - & 71.6 & - \\
    \midrule[1pt]
    \multicolumn{6}{l}{\textit{Training Algorithms}} \\
    \midrule[1pt]
    LLaMa2-7B & AgentGym & 76.6 & - & 63.3 & - \\ 
    Qwen2.5-7B & SFT   & 63.3 & - & 57.0 & - \\ 
    Qwen2.5-7B & GiGPO & \textcolor{grayrlvmr}{89.5} & - & \textcolor{grayrlvmr}{90.2} & - \\ 
    Qwen2.5-7B & RLVMR & \textcolor{grayrlvmr}{\underline{91.4}} & - & \textcolor{grayrlvmr}{\textbf{91.8}} & - \\ 
    Qwen2.5-3B & GRPO &  87.5 & 63.3 & 83.3 & 63.5 \\
    Qwen2.5-3B & EPO & \textbf{91.7} & \underline{75.8} & \underline{89.6} & \underline{75.4} \\
    \rowcolor{blue!10} Qwen2.5-3B & $\chi$PO & \textbf{91.7} & \textbf{78.2} & \textbf{91.8} & \textbf{75.9} \\
    \bottomrule[1.5pt]
    \end{tabular}
\end{table}

\begin{figure*}[t]
    \centering
    \begin{subfigure}{0.48\textwidth}
        \centering
        \includegraphics[width=\linewidth]{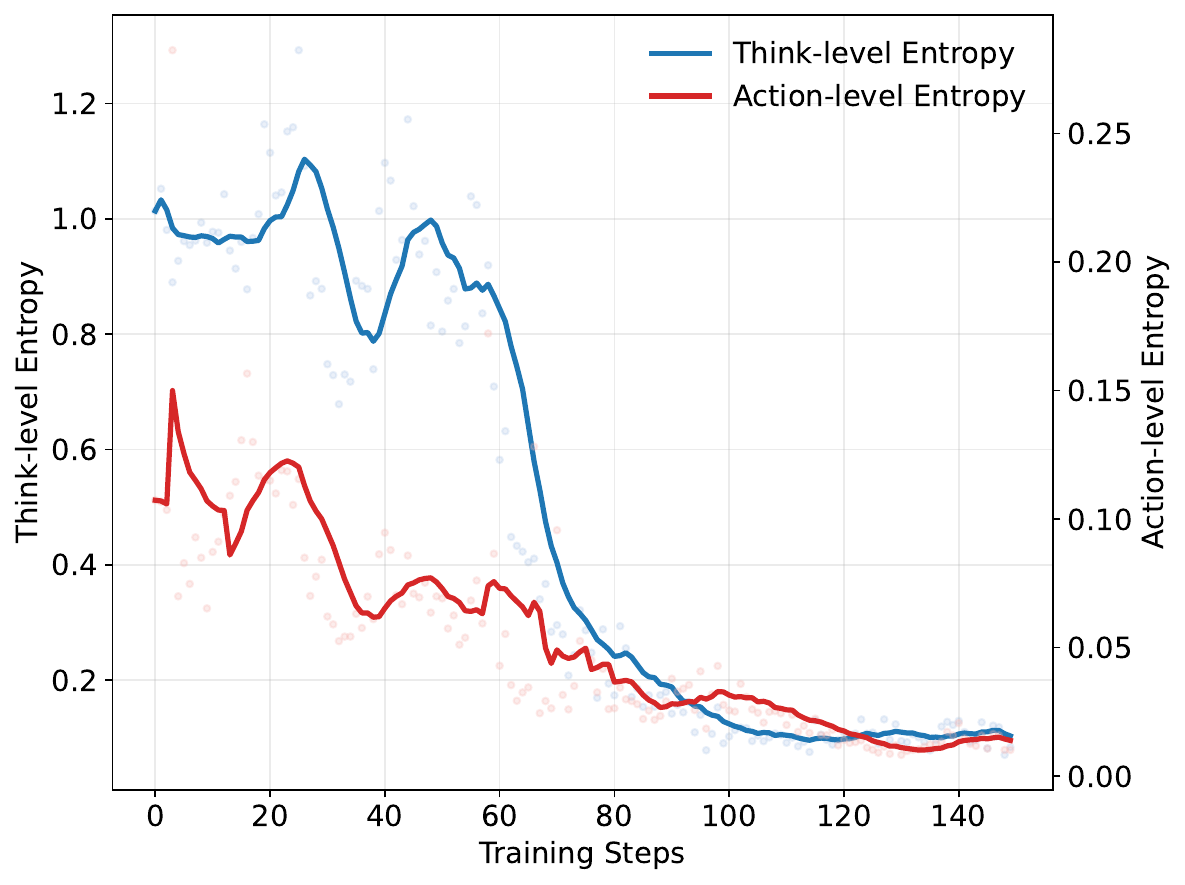}
        \caption{Think-level and action-level entropy in ALFWorld.}
        \label{fig:entropy}
    \end{subfigure}
    \hfill
    \begin{subfigure}{0.48\textwidth}
        \centering
        \includegraphics[width=\linewidth]{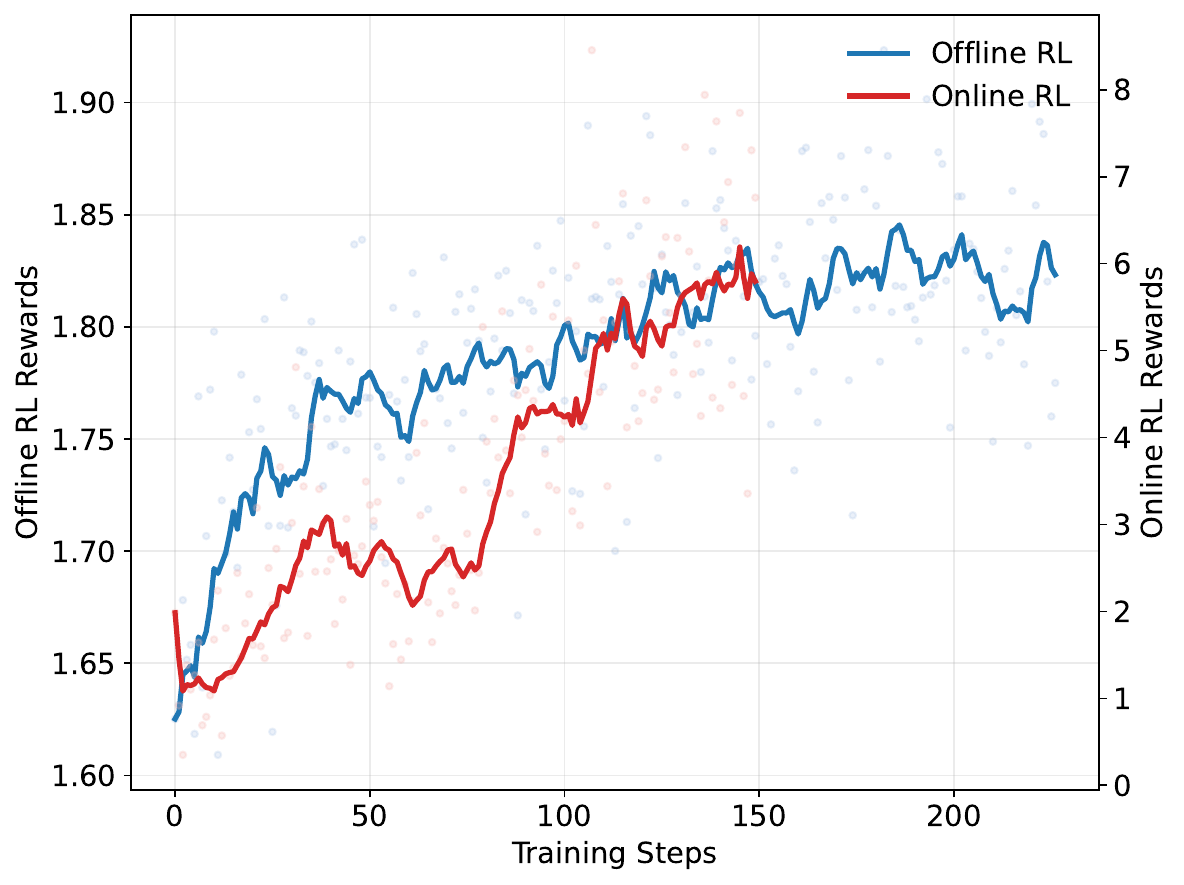}
        \caption{Online and offline RL rewards.}
        \label{fig:rewards}
    \end{subfigure}
    \caption{RL training dynamics analysis. (a) The entropy at the think level is approximately an order of magnitude higher than that at the action level. (b) Both online and offline RL exhibit steadily increasing reward accumulation over the course of training.}
\end{figure*}

This section presents a comprehensive evaluation of the proposed $\chi$PO in the online environment ALFWorld \citep{ALFWORLD}. ALFWorld comprises 4,639 interactive household task instances spanning six categories, including \texttt{Pick \& Place}, \texttt{Examine in Light}, \texttt{Clean \& Place}, \texttt{Heat \& Place}, \texttt{Cool \& Place}, \texttt{Pick Two \& Place}. These tasks require long-horizon planning, sequential decision-making, and robust spatial reasoning. For each category, task instances are organized into a \texttt{train}split, along with separate \texttt{seen} and \texttt{unseen} validation splits used for evaluation. The \texttt{seen} split contains tasks executed in environments encountered during training but with altered object configurations, counts, and visual properties. In contrast, the \texttt{unseen} split consists of entirely new task instances situated in previously unobserved rooms with novel receptacles and layout structures.

\paragraph{Experiment Setting.} Our evaluation follows the protocol introduced in EPO \citep{EPO}, reporting performance under both \textbf{IID} (in-distribution) conditions on the \texttt{seen} split and \textbf{OOD} (out-of-distribution) conditions on the \texttt{unseen} split. To ensure robustness and reproducibility, all experiments are conducted using five random seeds. We report two success metrics: $\textbf{Succ.}^*$ , defined as the mean of the best success rates achieved across the five random seeds, and $\overline{\textbf{Succ.}}$, which reflects the average steady-state performance after convergence over the same set of runs. More information about the experimental settings are available in Appendix \ref{evaluationcode}.

\paragraph{Baselines.} We adopt baseline results reported in EPO, encompassing prompt-based approaches such as ReAct \citep{react}, which rely exclusively on in-context reasoning without parameter updates. We further compare against Supervised Fine-Tuning (SFT) and general-purpose reinforcement learning methods, including GRPO \citep{GRPO}. In addition, we evaluate approaches specifically designed for these embodied or interactive agents in the online environment, including GiGPO \citep{gigpo}, RLVMR \citep{rlvmr}, and EPO \citep{EPO}. Following the experimental setup of EPO, we use the Qwen2.5 models rather than the Qwen3 series in this experiment.

\paragraph{Results.} As shown in Table~\ref{tab:onlinerl_result}, we compare our MagicAgent model with other ultra-scale and large-scale models under a ReAct-based framework. Specifically, MagicAgent-32B achieves the best performance, with \textbf{81.4\% IID} and \textbf{78.4\% OOD} success rates, substantially outperforming the representative ultra-scale state-of-the-art GPT-5.2 (IID: $63.5\%$, OOD: $62.6\%$) and DeepSeek (IID: $68.8\%$, OOD: $76.1\%$) models.

\paragraph{Effectiveness of $\chi$PO.} In addition, we compare multiple training algorithms to verify the effectiveness of the proposed $\chi$PO algorithm. As shown in Table~\ref{tab:onlinerl_result}, $\chi$PO achieves the strongest overall performance, attaining both the highest peak success rates (\textbf{IID: 91.7\%, OOD: 91.8\%}) and best converged success rates \textbf{(IID: 78.2\%, OOD: 75.9\%}), outperforming the EPO method. Moreover, $\chi$PO with the Qwen2.5-3B model achieves performance comparable to GiGPO and RLVMR using the larger Qwen2.5-7B model, indicating that the proposed exploration-exploitation mechanisms improve learning efficiency rather than relying solely on model scale. To further examine the think-level and action-level entropy smoothing mechanisms, Figure \ref{fig:entropy} shows that entropy at the think level is approximately an order of magnitude higher than that at the action level, and that both decrease throughout training. In addition, rewards in both offline and online reinforcement learning exhibit consistent upward trends in Figure \ref{fig:rewards}. These consistent improvements indicate that $\chi$PO not only accelerates early exploration to discover high-reward behaviors, but also stabilizes long-term performance through more reliable action execution. This behavior aligns with our design choices: token-level entropy regularization, think-level and action-level entropy smoothing promote broader exploration, which is particularly beneficial under sparse rewards, while the information bottleneck suppress unnecessary uncertainty at convergence, resulting in higher steady-state success rates. 

\subsection{Performance of MoE}
\label{MOE-exp}
\begin{table}[!h]
\centering
\caption{Ablation Study of Joint Optimization Strategy for MoE Supervised Fine-Tuning. \textbf{Bold} indicates the best performance in each benchmark, while \underline{Underline} indicates the second-best performance.}
\vspace{.1cm}
\renewcommand\arraystretch{1.15}  
\resizebox{0.8\textwidth}{!}{
\begin{tabular}{lcccccccc}
\toprule[1.5pt]
\multirow{2}{*}{\textbf{Method}} & \multicolumn{2}{c}{\textbf{WorfBench}} & \multicolumn{3}{c}{\textbf{NaturalPlan}} & \multicolumn{2}{c}{\textbf{$\boldsymbol{\tau^2}$-Bench}} \\
& {$F_1$ Chain} & {$F_1$ Graph} & {Trip} & {Meeting} & {Calendar} & {Retail} & {Airline} \\
\midrule[1pt]
BP & 79.0 & 68.4 & 39.5 & \underline{59.5} & 54.5 & 20.8 & 39.5 \\
BP + $\mathrm{LBL}_{\mathrm{micro}}$ & 78.5 & 69.0 & 38.1 & 57.8 & 52.4 & 43.9 & 47.0 \\
BP + VarianceLoss & 78.5 & 68.5 & 39.8 & 57.4 & 53.9 & 28.9 & 40.0 \\
BP + $\mathrm{LBL}_{\mathrm{seq}}$ & 77.9 & 68.4 & 38.1 & 55.5 & 54.5 & 26.1 & 41.0 \\
BP + $\mathrm{LBL}_{\mathrm{gbl}}$ & \textbf{79.3} & \textbf{70.2} & \underline{40.0} & 58.5 & 53.2 & 45.0 & 44.0 \\[3pt]
BP + z\_loss & 78.3 & 68.2 & 38.3 & 57.5 & 53.6 & \underline{51.3} & \underline{48.0} \\
BP + Expert Capacity & 77.9 & 67.6 & 39.6 & 57.6 &  \textbf{55.8} & 37.5 & 46.0 \\
\rowcolor{blue!10} BP + $\mathrm{LBL}_{\mathrm{gbl}}$ + z\_loss & \underline{79.0} & \underline{69.2} & \textbf{41.8} & \textbf{60.3} & \underline{54.9} & \textbf{52.6} & \textbf{52.5} \\
\bottomrule[1.5pt]
\end{tabular}}
\label{tab:baseline}
\end{table}

\paragraph{Joint Optimization Strategies for Enhancing Model Performance.} We compare the effects of applying different training tricks individually with a joint optimization strategy during the MoE SFT stage. The methods under comparison mainly include z-loss, micro-batch LBL, sequence-level LBL, global-batch LBL, variance loss, and expert capacity control. We perform ablation studies using the same training data, which includes multi-constraint scheduling, procedural logic orchestration, and long-horizon tool execution. With identical hyperparameters, the joint optimization strategy achieves superior benchmark metrics.

In our training data, Long-Horizon Tool Execution task exhibits both high complexity and long context, constituting the primary challenge for mixed training. Under this condition, most mainstream training tricks fail to maintain balanced performance. As shown in Table \ref{tab:baseline}, variance loss and sequence-level LBL exhibit suboptimal performance on $\tau^2$-Bench: retail and airline scores are mostly below 45. In contrast, z-loss performs clearly better, with 51.3\% in retail and 48.0\% in airline. By penalizing extreme routing logits, z-loss mitigates overly sharp routing distributions, yields smoother gradients, and stabilizes training. This is crucial for long-horizon tool execution training, where long sequences increase the numerical sensitivity of MoE.

Furthermore, global-batch LBL contributes by adopting a global optimization perspective to realize corpus-level load balancing and enhance expert specialization. On WorfBench, 
it achieves the highest scores on both F1 Chain (79.3\%) and F1 Graph (70.2\%), indicating more principled routing and improved expert allocation.

Finally, combining z-loss with global-batch LBL achieves balanced performance: on $\tau^2$-Bench, retail reaches 52.6\% and airline 52.5\%, with consistent gains across other benchmarks. These results suggest that jointly z-loss and global load balancing loss is an effective strategy for mixed training.

\paragraph{Balanced Expert Loading and Specialization.} We report the $\mathrm{MaxVio}_{\mathrm{batch}}$ value during training to evaluate the impact of different strategies on training load balancing and efficiency. $\mathrm{MaxVio}_{\mathrm{batch}}$ is defined as:
\begin{align}
\mathrm{MaxVio_{batch}} = \frac{\max_{i} \mathrm{Load}_i - \overline{\mathrm{Load}_i}}{\overline{\mathrm{Load}_i}},
\end{align}
where $\mathrm{Load}_i$ denotes the number of tokens assigned to the $i$-th expert per training batch, and $\overline{\mathrm{Load}_i}$ represents the ideal load distributed under perfect load balance.

\begin{figure*}[t]
    \centering
    \begin{subfigure}{0.48\textwidth}
        \centering
        \includegraphics[width=\linewidth]{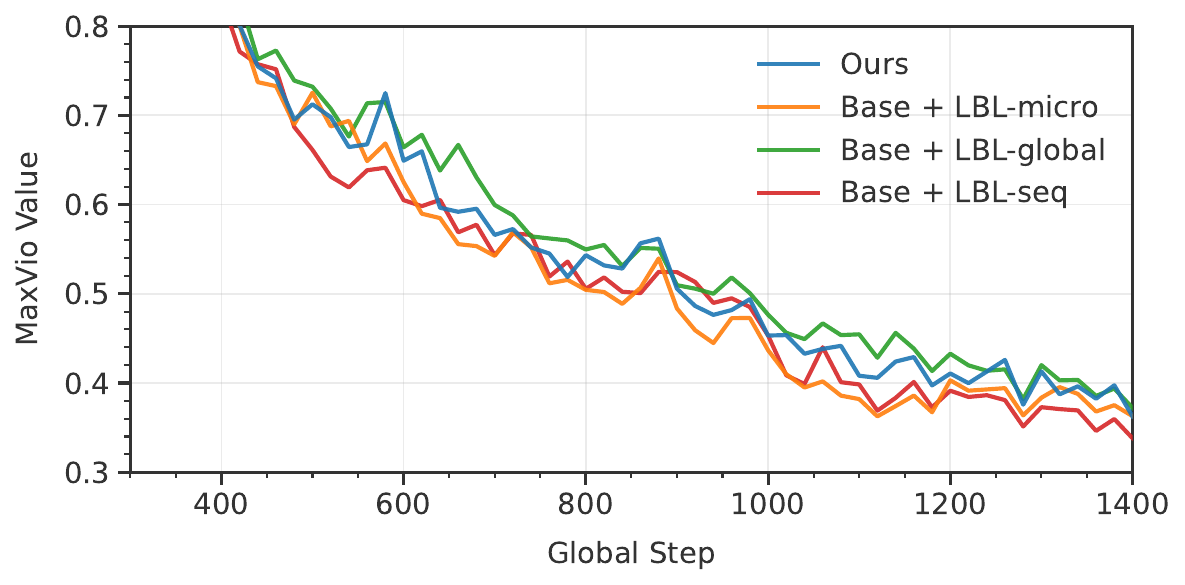}
        \caption{MaxVio curve (Lower range).}
        \label{fig:MaxVio curve (Lowerr range).}
    \end{subfigure}
    \hfill
    \begin{subfigure}{0.48\textwidth}
        \centering
        \includegraphics[width=\linewidth]{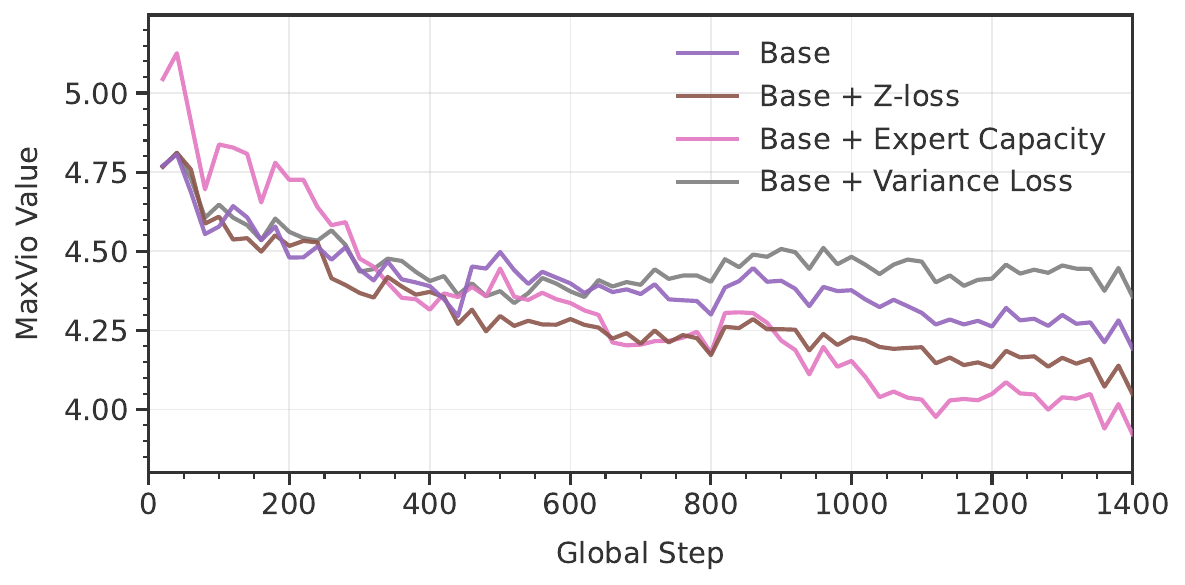}
        \caption{MaxVio curve (Higher range).}
        \label{fig:MaxVio curve (Higher range).}
    \end{subfigure}
    \caption{MaxVio value under various load-balanced strategies. (a) illustrates the MaxVio curves for $\mathrm{LBL}_{\mathrm{micro}}$, $\mathrm{LBL}_{\mathrm{gbl}}$, $\mathrm{LBL}_{\mathrm{seq}}$ and joint optimization strategy. (b) presents the MaxVio metrics for the baseline, z-loss, Expert Capacity, and Variance Loss methods.}
    \label{fig:RL}
\end{figure*}

\begin{figure*}[t]
    \centering
    \begin{subfigure}{0.32\textwidth}
        \centering
        \includegraphics[width=\linewidth]{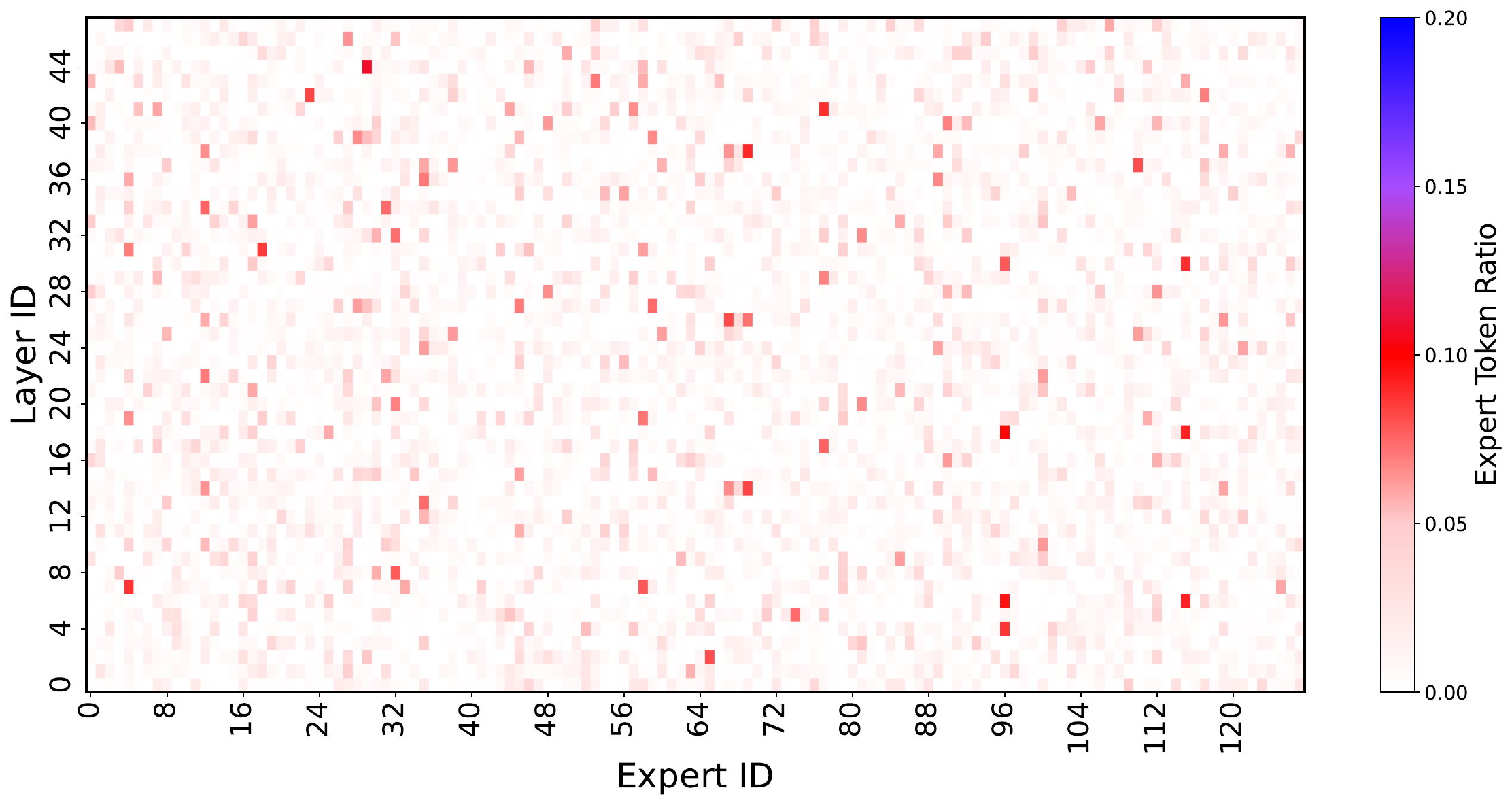}
        \caption{Calendar task.}
        \label{fig:heatmap_calendar}
    \end{subfigure}
    \hfill
    \begin{subfigure}{0.32\textwidth}
        \centering
        \includegraphics[width=\linewidth]{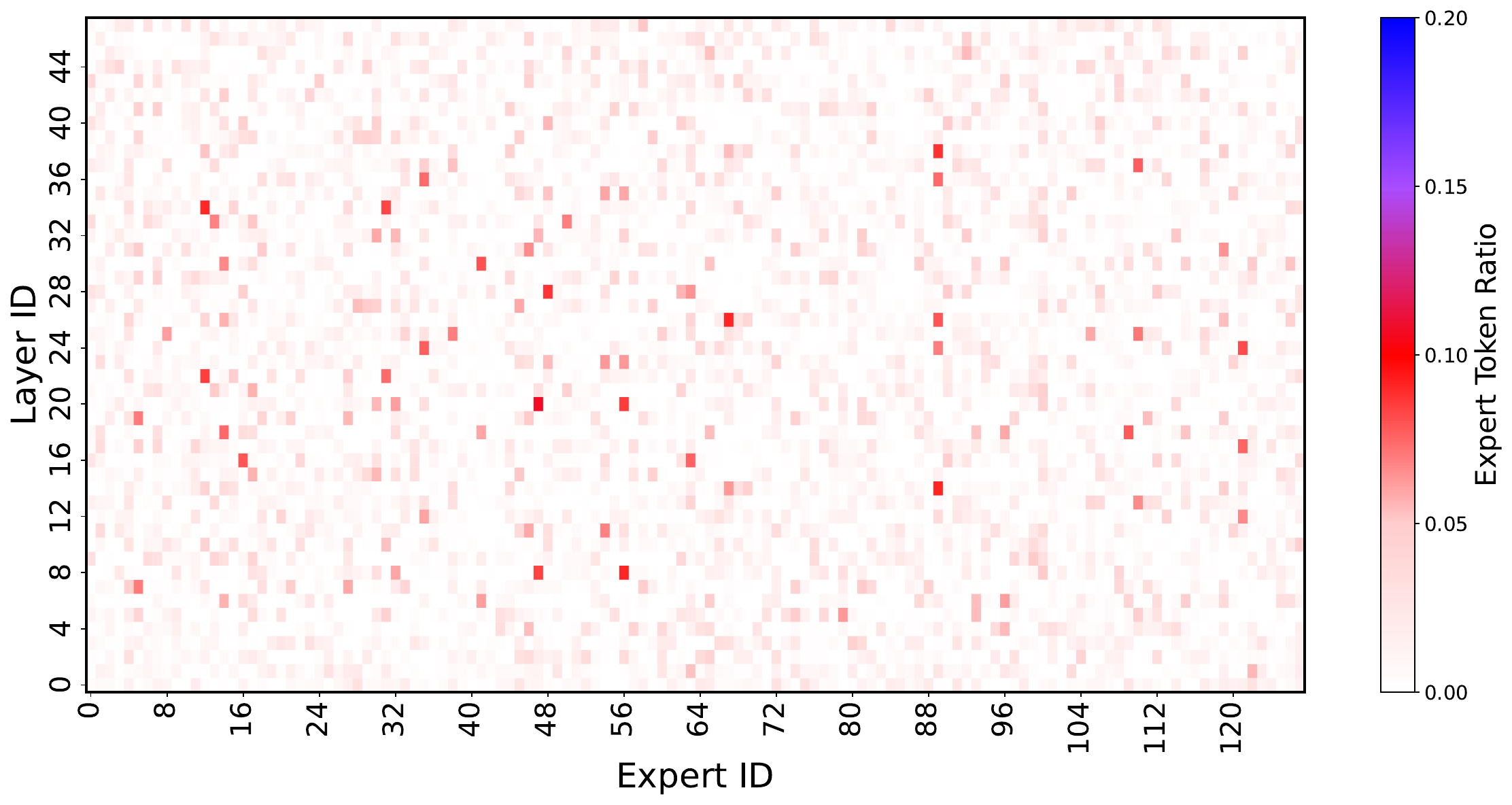}
        \caption{Meeting task.}
        \label{fig:heatmap_meeting}
    \end{subfigure}
    \hfill
    \begin{subfigure}{0.32\textwidth}
        \centering
        \includegraphics[width=\linewidth]{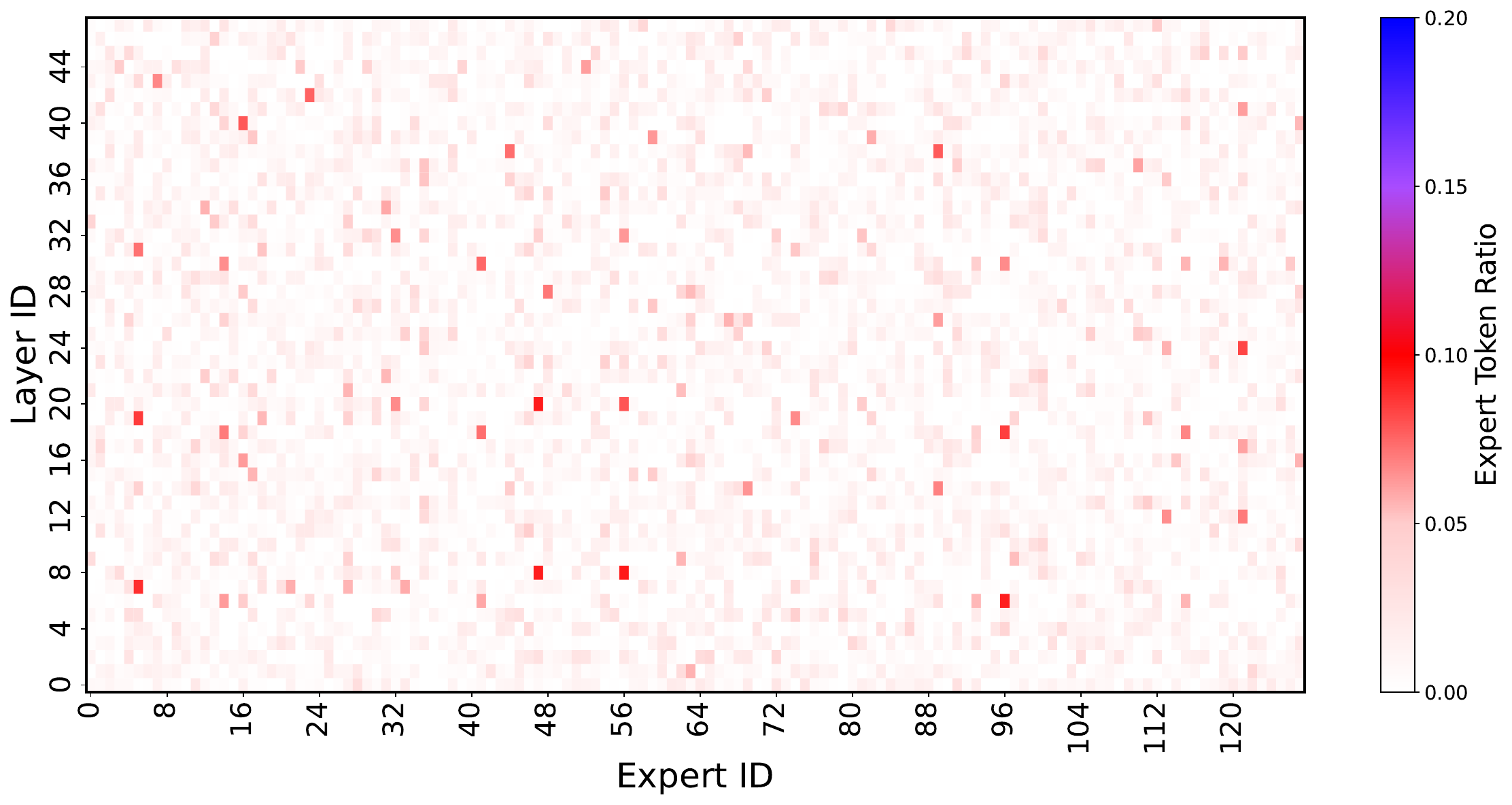}
        \caption{Trip task.}
        \label{fig:heatmap_trip}
    \end{subfigure}

    \caption{Expert frequency heatmaps under different agent tasks on NaturalPlan. (a) Calendar task, (b) Meeting task, and (c) Trip task. The distinct expert utilization patterns across tasks indicate the emergence of stable task-specific expert specialization during training.}
    \label{fig:overall_heatmap}
\end{figure*}

As shown in Figure \ref{fig:MaxVio curve (Higher range).}, both the expert capacity constraint and the z-loss improve training load balance over the baseline to a certain extent. While the variance loss encourages more discriminative routing decisions to enhance model performance, the absence of explicit load balancing constraints leads to a slight expert load imbalance compared to the baseline. As illustrated in Figure \ref{fig:MaxVio curve (Lowerr range).}, the joint optimization strategy achieves a balance in training load comparable to that of mainstream load balancing losses, while substantially outperforming the baseline and other competing approaches. This strategy ensures a significantly more stable and equitable distribution of training loads throughout the entire optimization trajectory.

Furthermore, we visualize the expert selection frequencies as heatmaps to analyze expert specialization patterns under different strategies. As illustrated in Figure \ref{fig:overall_heatmap}, distinct routing allocation patterns emerge across the three NaturalPlan subtasks: Trip, Meeting, and Calendar. This observation indicates that the model gradually develops stable expert specialization during training, enabling it to adaptively activate task-relevant experts to effectively handle different tasks.

\begin{table}[!t]
\centering
\caption{Inference Efficiency Comparison between Dense and MoE Models across Various Concurrency Levels.}
\vspace{.1cm}
\label{tab:model_performance}
\begin{tabular}{l c c c c}
\toprule[1.5pt]
\textbf{Model} & \textbf{Avg TTFT (s)} & \textbf{Avg TPOT (s)} & \textbf{Avg Latency (s)} & \textbf{Concurrency} \\
\midrule[1pt]
MagicAgent-32B      & 0.498 & 0.016 & 16.722 & 1 \\
MagicAgent-30B-A3B & 0.341(-31.5\%) & 0.007(-56.3\%) & 7.582(-54.7\%) & 1 \\
\midrule 
MagicAgent-32B      & 0.785 & 0.018 & 19.586 & 10 \\
MagicAgent-30B-A3B & 0.558(-28.9\%) & 0.011(-38.9\%) & 11.783(-39.9\%) & 10 \\
\midrule 
MagicAgent-32B      & 2.851 & 0.024 & 27.855 & 50 \\
MagicAgent-30B-A3B & 1.153(-59.6\%) & 0.016(-33.3\%) & 17.891(-35.8\%) & 50 \\
\midrule 
MagicAgent-32B      & 4.009 & 0.035 & 39.555 & 100 \\
MagicAgent-30B-A3B & 2.250(-43.9\%) & 0.021(-40.0\%) & 23.478(-40.6\%) & 100 \\
\bottomrule[1.5pt]
\end{tabular}
\end{table}

\paragraph{Inference Efficiency.} In addition, we evaluate the inference efficiency\footnote{The evaluation is based on the EvalScope framework.} of our proposed models, specifically comparing the Dense model (MagicAgent-32B) and the MoE model (MagicAgent-30B-A3B). The benchmarks are conducted on a server equipped with 4 NVIDIA A800 GPUs, using a standardized configuration of 1024 input tokens and 1024 output tokens. As illustrated in Table \ref{tab:model_performance}, the MoE architecture demonstrates a substantial performance advantage over the Dense counterpart across all measured metrics. Notably, our MoE model achieves a reduction of 28.9\% to 59.6\% in Average TTFT and 33.3\% to 56.3\% in Average TPOT across various concurrency levels. These results underscore that our MoE model significantly reduces computational overhead and overall latency (by up to 54.7\%) while maintaining high throughput, making it highly suitable for large-scale deployment.

\section{Conclusions}
This paper presents \textbf{MagicAgent}, a model series that addresses the bottlenecks of ``single-point optimizations'' and balances multi-type agent planning tasks in a unified framework. To mitigate the scarcity of high-quality and diverse agentic data, a systematic yet lightweight framework for agentic data synthesis is purposed, covering tasks across hierarchical task decomposition, tool-augmented planning, multi-constraint scheduling, procedural logic orchestration and long-horizon tool execution. Moreover, we develop a two-stage training framework centered on multi-task reinforcement learning(RL), combined with an online RL method named $\chi$PO that balances exploration and exploitation. Empirical validation across public and proprietary benchmarks demonstrates that MagicAgent achieves superior performance among both Ultra-Scale Models and Large-Scale Models. Driven by its strong capabilities in planning, the MagicAgent model series has been deployed across multiple high-value scenarios under Honor's intelligent assistant. We present three representative application domains to illustrate the versatility and efficacy of our approach in Appendix \ref{product_usage}. In future work, our next step is to push beyond current model limits by advancing long-horizon reasoning and code-augmented task completion, while leveraging memory and personalized information to unlock new frontiers in agent intelligence.

\bibliographystyle{delta_tuning}  
\bibliography{references}  


\clearpage
\appendix
\raggedbottom

\section*{Appendix}
\section{Training Details}
\label{training_details}

We provide a comprehensive overview of the training hyperparameters for MagicAgent-32B and MagicAgent-30B-A3B-Instruct-2507, including core model configurations, mixture-of-experts (MoE) parallelization strategies, and optimization settings. Table \ref{tab:training_detail} details prompt and generation limits, optimizer and scheduler choices, and numerical precision, along with parallelism parameters tailored for large-scale MoE training. It further specifies the hyperparameters used for both Supervised Fine-tuning (SFT) and Reinforcement Learning (RL), including batch sizes, learning rates, generation strategies, and reward weighting, ensuring reproducibility and stable optimization across training stages.

\begin{table}[!h]
\centering
\caption{Training parameters for the MagicAgent-32B and MagicAgent-30B-A3B models.}
\vspace{.2cm}
\label{tab:training_detail}
\renewcommand{\arraystretch}{1.2}
\begin{tabular}{lcl}
\toprule[1.5pt]
\textbf{Parameter} & \textbf{Default Value} & \textbf{Description} \\
\midrule[1pt]
max\_length & 32768 & Maximum prompt length and completion length \\
num\_train\_epochs & 1 & Number of training epochs \\
weight\_decay & 0.01 & Weight decay coefficient \\
adam\_beta1 & 0.9 & Adam optimizer beta1 parameter \\
adam\_beta2 & 0.95 & Adam optimizer beta2 parameter \\
lr\_scheduler\_type & cosine & Learning rate scheduler type \\
bf16 & True & Use bfloat16 precision \\
\midrule[1pt]
\multicolumn{3}{l}{\textbf{MoE Setting}} \\
tensor\_model\_parallel\_size & 4 & Tensor parallel size \\
pipeline\_model\_parallel\_size & 2 & Pipeline parallel size \\
context\_parallel\_size & 2  & Context parallel size \\
expert\_model\_parallel\_size & 4 & Expert parallel size \\
moe\_aux\_loss\_coeff & 1e-4  & Global aux loss coefficient \\ 
moe\_z\_loss\_coeff & 5e-4  & Z-Loss coefficient \\
\midrule[1pt]
\multicolumn{3}{l}{\textbf{Supervised Fine-tuning (SFT)}} \\
num\_gpus & 32 & Number of GPUs (4 machines $\times$ 8 GPUs) \\
per\_device\_train\_batch\_size & 4 & Training batch size per device \\
gradient\_accumulation\_steps & 1 & Gradient accumulation steps \\
learning\_rate & 1e-5 & Learning rate of SFT \\
\midrule[1pt]
\multicolumn{3}{l}{\textbf{Reinforcement Learning (RL)}} \\
num\_gpus & 32 & Number of GPUs (4 machines $\times$ 8 GPUs) \\
per\_device\_train\_batch\_size & 1 & Training batch size per device \\
gradient\_accumulation\_steps & 1 & Gradient accumulation steps \\
learning\_rate & 1e-6 & Learning rate of actor policy \\
num\_generations & 8 & Number of generations in RL \\
temperature & 0.9 & Temperature parameter for generations in RL \\
top\_p & 0.96 & Cumulative probability threshold for generation \\
top\_k & 20 & Number of top tokens filtered by highest probability \\
$\alpha_1,\alpha_2,\alpha_3,\alpha_4,\alpha_5$ & 1.0 & Weighting coefficients of the respective reward components \\
\bottomrule[1.5pt]
\end{tabular}
\end{table}
\section{Evaluation Settings}
\label{evaluationcode}
All evaluations were conducted in strict adherence to the protocols outlined in the official public repositories. To ensure a fair comparison, all models were evaluated in standard inference mode (i.e., with any extended \textit{reasoning} or \textit{thinking} features disabled). The evaluation parameters were set as the recommended values. The specific codebases utilized for benchmarking are listed below:

\begin{itemize}
    \item \textbf{NaturalPlan}: \href{https://github.com/google-deepmind/natural-plan}{google-deepmind/natural-plan}

    \item \textbf{Worfbench}: \href{https://github.com/zjunlp/WorfBench}{zjunlp/WorfBench}

    \item \textbf{BFCL-v3}: \href{https://github.com/ShishirPatil/gorilla}{ShishirPatil/gorilla} \footnote{Given the rapid iteration of this repository, we utilized the specific commit \href{https://github.com/ShishirPatil/gorilla/commit/cd9429ccf3d4d04156affe883c495b3b047e6b64}{\texttt{cd9429}} to ensure reproducibility.}

    \item \textbf{ACEBench}: \href{https://github.com/chenchen0103/ACEBench}{chenchen0103/ACEBench}

    \item \textbf{$\boldsymbol{\tau^2}$-Bench}: \href{https://github.com/sierra-research/tau2-bench}{sierra-research/tau2-bench}

\end{itemize}

We observed that certain model families, such as the GLM series, have introduced an optimized user simulator to enhance evaluation performance on the ${\tau^2}$-Bench. However, as the implementation details and source code for this specialized simulator have not been publicly released, we adhere strictly to the official evaluation protocols and codebase to ensure a standardized and fair comparison across all models.

Evaluating on the ALFWorld benchmark necessitates a critical comprehension of complex environmental contexts. Consequently, relying on a \textit{standard inference mode} is inadequate for this evaluation. To ensure a fair comparison among different models within a unified sandbox environment, we selected a subset of baseline models recognized for their robust reasoning capabilities. For the ALFWorld benchmark, the selected models were evaluated in reasoning inference mode (i.e., with extended \textit{reasoning} or \textit{thinking} features enabled). The evaluation sandbox strictly adheres to the official open-source codebase:

\textbf{ALFWorld}: \href{https://github.com/langfengQ/verl-agent}{verl-agent}

\newtcolorbox{promptbox}[1]{
    colback=gray!10,        
    colframe=black,         
    boxrule=1pt,            
    arc=5pt,                
    left=0pt, right=0pt,    
    top=0pt, bottom=5pt,    
    enhanced,               
    title=#1,
    fonttitle=\bfseries\sffamily\Large,
    coltitle=white,
    colbacktitle=black,     
    attach boxed title to top left={xshift=0mm, yshift=0mm},
    boxed title style={sharp corners, size=small, frame hidden}
}

\newcommand{\sectionbar}[1]{
    \begin{tcolorbox}[
        colback=black!85, 
        colframe=black!85,
        sharp corners,
        left=6pt,
        right=6pt,
        top=3pt,
        bottom=2pt,
        grow to left by=0pt,
        grow to right by=0pt,
        boxrule=0pt
    ]
    \textcolor{white}{\sffamily\bfseries\large #1}
    \end{tcolorbox}
}

\lstset{
    basicstyle=\small\ttfamily,
    columns=flexible,
    breaklines=true,
    showstringspaces=false,
    xleftmargin=20pt,
    lineskip=2pt,
    upquote=true
}

\section{Hierarchical Task Decomposition Examples}
\label{td_examples}

\ExampleBox{Raw Format}{Appendix/hierarchical-task-decomposition1.txt}
\vspace{10pt}
\ExampleBox{Raw Format}{Appendix/hierarchical-task-decomposition2.txt}

\section{Tool-Augmented Planning Examples}
\label{tc_examples}
\ExampleBox{Raw Format}{Appendix/tool-augmented-planning.txt}

\section{Multi-Constraint Scheduling Examples}
\label{mcs_examples}
\ExampleBox{Raw Format}{Appendix/multi-constraint-scheduling.txt}

\section{Procedural Logic Orchestration Examples}
\label{plo_examples}
\ExampleBox{Raw Format}{Appendix/procedural-logic-orchestration.txt}

\section{Long-Horizon Tool Execution Examples}
\label{lhte_examples}
\ExampleBox{ShareGPT Format}{Appendix/long-horizon.txt}
\section{Usage}
\label{product_usage}

The proposed foundation model, MagicAgent, demonstrates broad applicability across a variety of practical and commercial scenarios. We outline three representative application domains to illustrate the versatility and efficacy of our approach.

\begin{figure}[ht]
    \centering
    \includegraphics[width=0.8\linewidth]{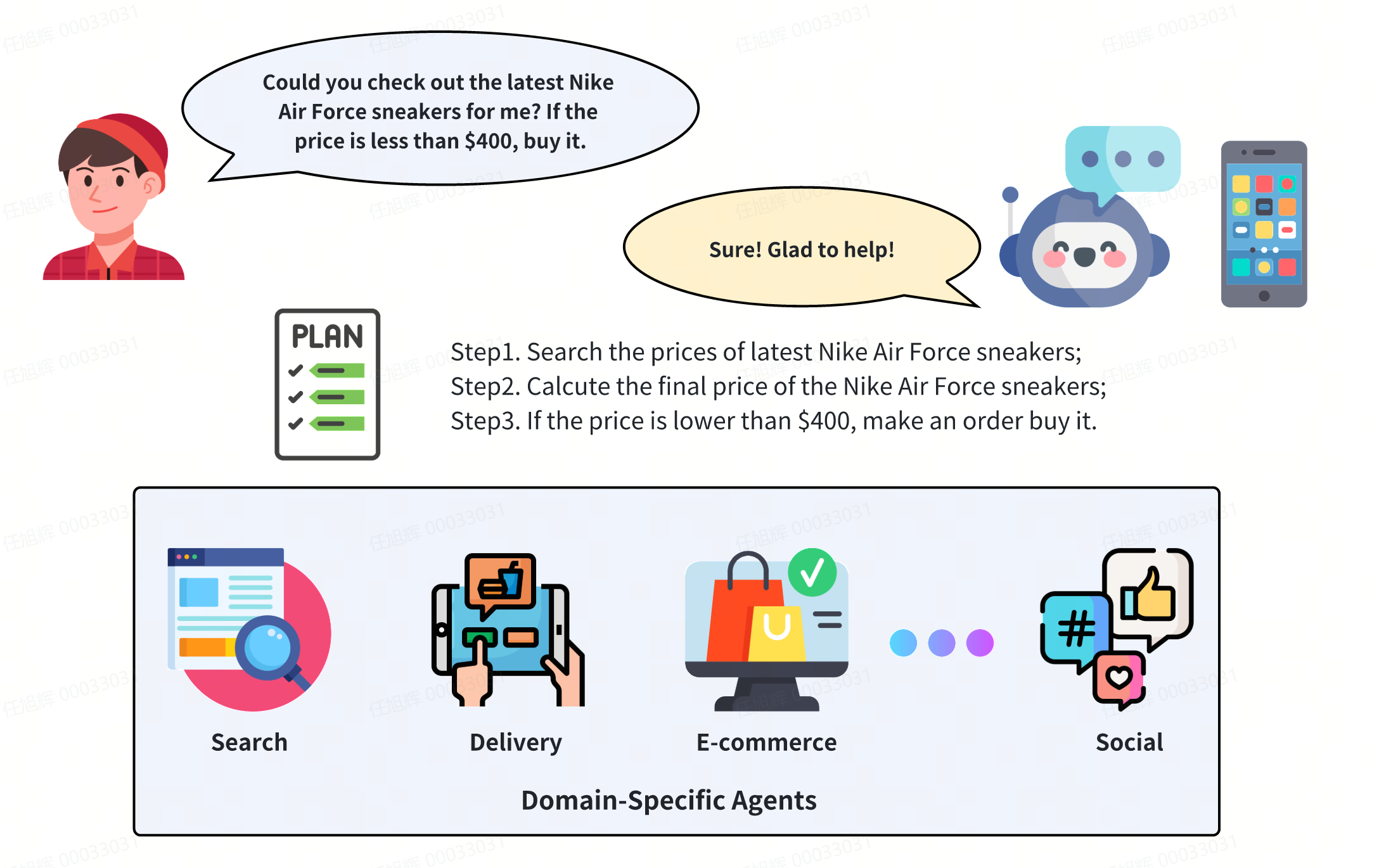}
    \caption{A demo case of Intent-Routing Agent in Multi-Agent architectures.}
    \label{fig:plehIntent-Routing}
\end{figure}

\begin{itemize}
    \item \textbf{Tool-calling Agent}. MagicAgent exhibits robust performance in tool-augmented applications. By effectively engaging with user queries, the model executes intelligent decisions regarding both tool selection and parameter instantiation—capabilities that serve as the fundamental backbone for most real-world, interactive agent deployments. The use case is shown in Figure \ref{fig:placTool-calling}.
    \begin{figure}[H]
        \centering
        \includegraphics[width=0.75\linewidth]{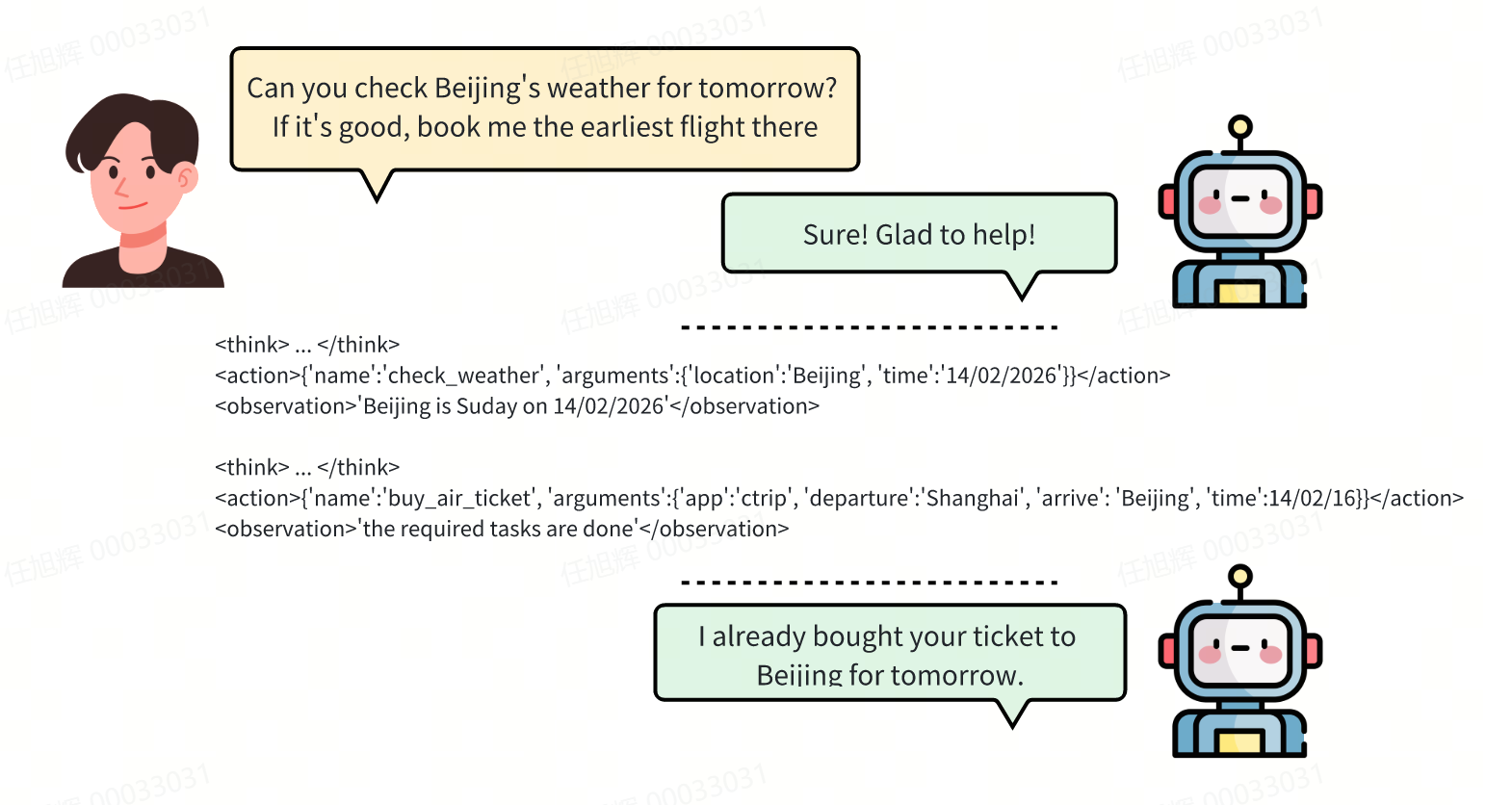}
        \caption{A demo case of Tool-calling Agent.}
        \label{fig:placTool-calling}
    \end{figure}

    \item \textbf{Travel Planning Agent}. Our model demonstrates exceptional proficiency in comprehending complex user requirements alongside dynamic environmental constraints. Leveraging this contextual information, MagicAgent functions as an intelligent travel assistant, rationally synthesizing and optimizing comprehensive travel itineraries. The use case is shown in Figure \ref{fig:placehTravel}.
    \begin{figure}[H]
        \centering
        \includegraphics[width=0.8\linewidth]{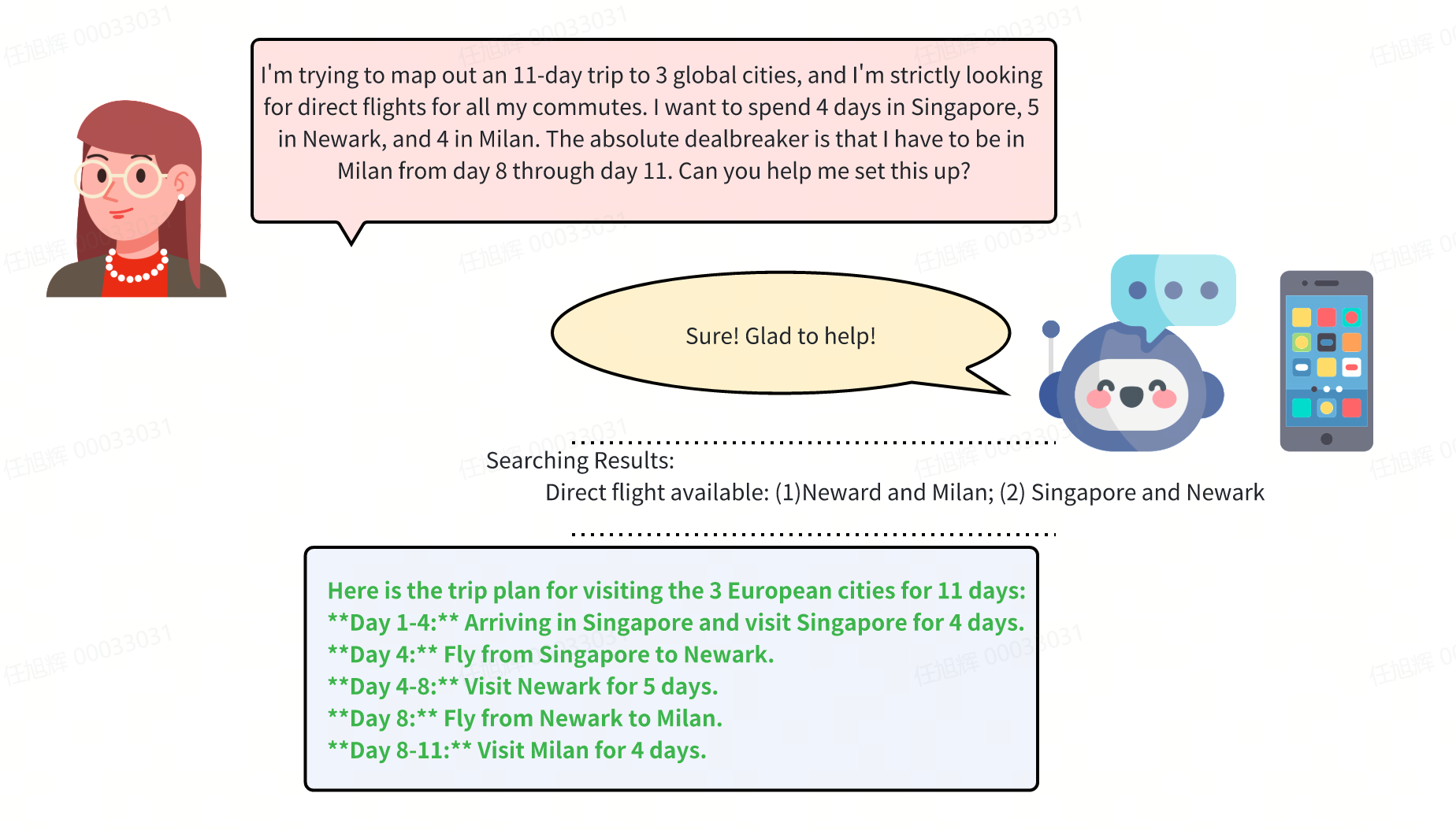}
        \caption{A demo case of Travel Planning Agent.}
        \label{fig:placehTravel}
    \end{figure}

    \item \textbf{Intent-Routing Agent in Multi-Agent Architectures}. MagicAgent is highly capable of decomposing complex, high-level user tasks into sequential, executable sub-tasks. Acting as a central intent router, it strategically orchestrates the deployment of downstream, domain-specific agents to achieve optimal and coherent outcomes. The use case is shown in Figure \ref{fig:plehIntent-Routing}.
\end{itemize}

\end{document}